\documentclass[journal]{elsarticle}
\usepackage{blindtext}
\usepackage{graphicx}
\usepackage{amsmath}
\usepackage{multirow}

\begin{document}
\begin{frontmatter}
\title{Relation Extraction : A Survey}
\author[a,b]{Sachin Pawar}
\ead{sachinpawar@cse.iitb.ac.in}
\author[a]{Girish K. Palshikar}
\author[b]{Pushpak Bhattacharyya}
\address[a]{TCS Research, Tata Consultancy Services Ltd.}
\address[b]{Department of CSE, Indian Institute of Technology Bombay}

%\maketitle

\begin{abstract}
%Occurrences of named entities in a sentence are often linked through well-defined relations; e.g., occurrences of person and organization in a sentence may be linked through relations such as {\em employed\_at}. Supervised techniques for Relation Extraction can be classified into two types - Kernel based methods and Feature based methods. Kernel based methods design kernel functions to compute similarities between representations of two relation instances (e.g. Shallow parse trees, dependency trees, dependency graph paths etc.) and employ Support Vector Machines for classification. Feature based methods represent each relation instance with a feature vector which captures various characteristics of the relation instance in terms of various lexical, syntactic and semantic features. SVMs or Maximum Entropy models are used as classifiers. A recent trend in Relation Extraction is Distant Supervision, where (possibly noisy) labelled data for relation extraction is created without any human efforts by taking help of some knowledge base of semantic relations (e.g. Freebase). Another recent trend is Open Information Extraction (OIE), which is a novel extraction paradigm that works without any pre-defined set of target relations. OIE facilitates domain-independent discovery of relations extracted from massive text corpora and it is scalable to the diversity and size of the web corpus.
%In today's world
With the advent of the Internet, large amount of digital text is generated everyday in the form of news articles, research publications, blogs, question answering forums and social media. It is important to develop techniques for extracting information automatically from these documents, as lot of important information is hidden within them. This extracted information can be used to improve access and management of knowledge hidden in large text corpora. Several applications such as Question Answering, Information Retrieval would benefit from this information. Entities like persons and organizations, form the most basic unit of the information. Occurrences of entities in a sentence are often linked through well-defined relations; e.g., occurrences of person and organization in a sentence may be linked through relations such as {\em employed\_at}. The task of Relation Extraction (RE) is to identify such relations automatically. 
%The techniques for RE can be broadly classified into three types - supervised, semi-supervised and unsupervised. 
In this paper, we survey several important supervised, semi-supervised and unsupervised RE techniques. 
We also cover the paradigms of Open Information Extraction (OIE) and Distant Supervision.
%We also cover the paradigm of Open Information Extraction (OIE), which is a novel extraction paradigm that works without any pre-defined set of target relations. 
%OIE facilitates domain-independent discovery of relations extracted from massive text corpora and it is scalable to the diversity and size of the web corpus. 
%Another popular paradigm in Relation Extraction is Distant Supervision, where (possibly noisy) labelled data for relation extraction is created without any human efforts by taking help of some knowledge base of semantic relations (e.g. Freebase). 
Finally, we describe some of the recent trends in the RE techniques and possible future research directions.
This survey would be useful for three kinds of readers - i) Newcomers in the field who want to quickly learn about RE; ii) Researchers who want to know how the various RE techniques evolved over time and what are possible future research directions and iii) Practitioners who just need to know which RE technique works best in various settings.
\end{abstract}

\begin{keyword}
Relation Extraction \sep Supervised Learning \sep Kernel Methods \sep Unsupervised Learning \sep Semi-supervised Learning \sep Open Information Extraction \sep Distant Supervision
\end{keyword}

\end{frontmatter}

\section{Introduction}~\label{sec:intro}

It is well-known that a lot of tacit and experiential knowledge is present in document repositories (e.g., reports, emails, resumes, papers, proposals, blogs etc.) that are created and maintained within an enterprise or across the Web. Extracting this knowledge, disseminating it when needed and reusing it to improve decision-making and operational efficiency of practical tasks are important. The essential goal of {\em information extraction (IE)}~\cite{Moe06},~\cite{sarawagi2008information} is to extract a specific kind of information from a given document repository and output it to a structured repository such as a relational table or an XML file. %Figure~\ref{fig1} shows two examples of IE from unstructured text. 
%\begin{figure}
%\centering
%\label{fig1}
%\includegraphics[height=6cm,width=\columnwidth]{fig1.jpg}
%\caption{Two examples of IE from unstructured documents.}
%\normalsize
%\end{figure}
IE is an important problem that involves natural language processing, computational linguistics and text mining. IE is an useful first step in a wide range of knowledge management systems. % (Table~\ref{table1}). 
In addition, IE is also useful in other tasks such as information retrieval, question-answering (e.g., to answer questions like {\small {\tt Where is the Taj Mahal?}}) and so forth.
%\begin{table}[h]\footnotesize
%%\small
%\begin{tabular}{|p{4.5cm}|p{7cm}|}
%\hline
%\textbf{Repository} & \textbf{Use-cases for extracted information}\\
%\hline
%Resumes & Matching jobs, finding experts~\cite{PSP12} \\
%\hline
%Financial news & Identify competitor events and business model~\cite{PDP08}\\
%\hline
%Agriculture plant pathology documents & Help-desk for farmers~\cite{PPP13} \\
%\hline
%Customer support tickets & Reduce cost/time/efforts spent in support \\
%\hline
%Vehicle maintenance records & Reduce warranty costs~\cite{KPCD05} \\
%\hline
%Insurance claims & Fraud detection, reduce claim processing time \\
%\hline
%Survey responses & Identify complaints, issues, suggestions~\cite{PDB09} \\
%\hline
%Software requirements specifications & Detect errors / issues \\
%\hline
%Blogs and web-sites & Detect sentiments about product features \\
%\hline
%Emails for customer support & Identify complaints, automated reply \\
%\hline
%\end{tabular}
%\caption{Document repositories and use-cases for extracted information.}
%\label{table1}
%%\normalsize
%\end{table}

Information that users want to extract from documents is often of the following 3 kinds: (1) named entities, (2) relations and (3) events. In this paper, the focus is on Relation Extraction (RE). 
A {\em named entity (NE)} is often a word or phrase that represents a specific real-world object. As an example, {\small {\tt Barack Obama}} is a NE, and it has one specific {\em mention} in the following sentence: {\small {\tt Barack Obama is visiting India in January, 2015.}}. A {\em NE mention} in a particular sentence can be using the name itself ({\small {\tt Barack Obama}}), nominal ({\small {\tt US President}}), or pronominal ({\small {\tt he}}). NEs are often categorized into various {\em generic NE types}: PERSON, ORGANIZATION, LOCATION, DATE, TIME, PHONE, ZIPCODE, EMAIL, URL, AMOUNT etc. Other generic NEs include: FILM-TITLE, BOOK-TITLE etc. In {\em fine-gained NER}, the problem is to identify generic NE which are hierarchically organized; e.g., PERSON may be sub-divided into POLITICIAN, SCIENTIST, SPORTSPERSON, FILMSTAR, MUSICIAN etc. {\em Domain-specific NE} consist of, for example, names of proteins, enzymes, organisms, genes, cells etc., in the biological domain. NE in the manufacturing domain are: MANUFACTURER, PRODUCT, BRAND-NAME, FEATURE etc. {\em Named entity recognition (NER)} is the task of identifying all the mentions (occurrences) of a particular NE type in the given documents. For example: 
%{\footnotesize {\tt [Sachin Tendulkar]$_{PERSON}$ has not yet scored a century at [Lord's]$_{LOCATION}$ in [London]$_{LOCATION}$.}}\\
%{\footnotesize {\tt [J. P. Morgan]$_{ORG}$ strengthens domestic treasury management offering in [Malasia]$_{LOCATION}$.}}\\
{\footnotesize {\tt In a strategic reshuffle at [Bank of America - Merrill Lynch]$_{ORG}$, [Atul Singh]$_{PERSON}$ has taken over as managing director of Global Wealth and Investment Management in [India]$_{LOCATION}$.}} 
NER is an important sub-problem in IE; see~\cite{Pal12,NS07} %~\cite{Pal12},~\cite{NS07} 
for surveys of techniques for NER. 

A {\em relation} usually denotes a well-defined (having a specific meaning) relationship between two or more NEs. Some examples of relations are the MEMBER-AFFILIATION relation between PERSON and ORG, HAS relation between PRODUCT and FEATURE, AUTHOR-OF relation between PERSON and BOOK-TITLE and so forth. Example:\\
{\footnotesize {\tt [Bill Gates]$_{PERSON}$ announced that [John Smith]$_{PERSON}$ will be [the chief scientist]$_{PERSON}$ of [Microsoft Corporation]$_{ORG}$.}}\\
{\footnotesize {\tt The [Epson WorkForce 840's]$_{PRODUCT}$ [500-page paper capacity]$_{FEATURE}$ is convenient for high-volume office printing, and you can stock two different types of paper in [a pair of size-adjustable trays]$_{FEATURE}$.}}\\
We focus on binary relations and assume that both the argument NE mentions that participate in a relation mention occur in the same sentence. Note also that a relation need not exist between every pair of NE mentions in the given sentence. Example: 
{\footnotesize {\tt [John Smith]$_{PERSON}$ had visited [Bank of America]$_{ORG}$ in August 2003.}}\\ The task of {\em relation extraction (RE)} consists of identifying mentions of the relations of interest in each sentence of the given documents. Relation extraction is a very useful next step in IE, after NER. 

Successful RE requires detecting both the argument mentions, along with their entity types chaining these mentions to their respective entities, 
determining the type of relation that holds between them. Relation extraction faces many challenges. First, there is a vast variety of possible relations, which vary from domain-to-domain. Non-binary relations present special challenges. Supervised machine learning techniques applied to RE face the usual difficulty of a lack of sufficient training data. The notion of a relation is inherently ambiguous and there is often an inherent ambiguity about what a relation ``means'', which is often reflected in high inter-annotator disagreements.
As the expression of a relation is largely language-dependent, it makes the task of RE also language-dependent. Most of the work that we survey is concerned with English, and extending these techniques to non-English languages is not always easy.  

\subsection{Datasets for Relation Extraction}\label{sec:datasets}
\begin{figure}
\centering
\label{fig2}
\includegraphics[scale=0.5]{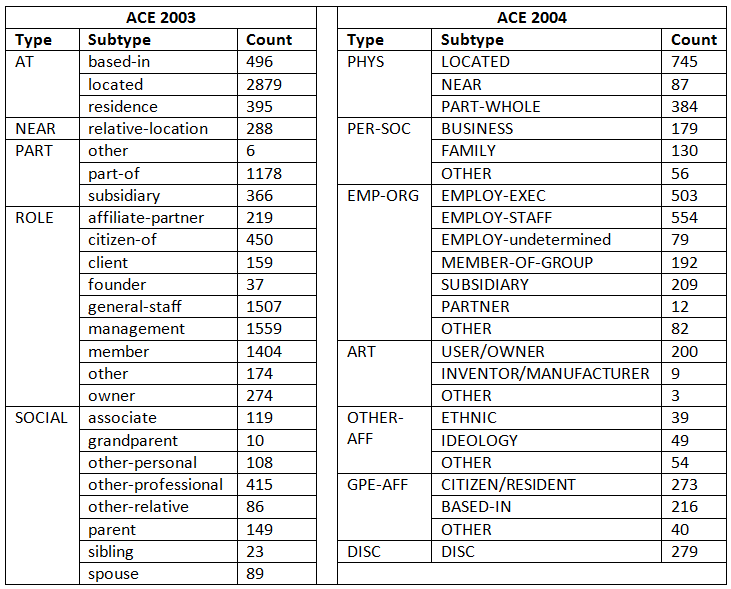}
%\caption{Relation Types and Sub-Types in ACE 2003 data, along with their occurrence counts ~\cite{Kam04}}.
\caption{Relation Types and Sub-Types in ACE 2003 and ACE 2004 datasets, along with their occurrence counts}.
\end{figure}

Automatic Content Extraction (ACE)~\cite{doddington2004automatic}\footnote{http://www.nist.gov/speech/tests/ace/}, \footnote{http://www.ldc.upenn.edu/Projects/ACE} is an evaluation conducted by NIST to measure the tasks of Entity Detection and Tracking (EDT) and Relation Detection and Characterization (RDC). ACE defines the following NE types: PERSON, ORG, LOCATION, FACILITY, GEO\_POLITICAL\_ENTITY (GPE), WEAPON etc. GPE refers to a geographically defined regions with a political boundary, e.g. countries, cities. The EDT task consists of detecting mentions of these NEs, and identifying their co-references. The RDC task consists of detecting relations between entities identified by the EDT task. The Figure~\ref{fig2} shows various relation types and subtypes defined in the ACE 2003~\cite{ace2003} and ACE 2004~\cite{ace2004} datasets. 
%ACE 2003 dataset~\cite{ace2003} defines 5 types of relations: AT, NEAR, PART, ROLE and SOCIAL. These relation subtypes are further sub-divided into specific relations (Figure~\ref{fig2}). The ACE corpus is labelled (entity and relation mentions are tagged in each sentence) and it is available from LDC (Linguistic Data Consortium). The training set consists of 674 labelled documents (~300k words) and 9683 instances of relations. ACE has only binary relations. Note that the order of mentions is usually significant in a  relation mention; e.g., P1 CITIZEN\_OF P2 is different from P2 CITIZEN\_OF P1.
ACE 2004 dataset~\cite{ace2004} is the most widely used dataset in the literature to report the performance of various relation extraction techniques. Some examples of the ACE 2004 relation types are shown in the Table~\ref{tab:ACE_relation_examples}.
%ACE 2004 revised the relation types (see table~\ref{tab:ACE_relation_examples}) defined earlier to come up to with set of 7 relation types : EMP\_ORG, GPE\_AFF, OTHER\_AFF, PER\_SOC, PHYS, ART, DISC. 
ACE released two more datasets, namely - ACE 2005~\cite{ace2005} and ACE 2007~\cite{ace2007}. Hachey et al.~\cite{hachey2012datasets} describes the ACE 2004 and ACE 2005 datasets and their standardization.
\begin{table}[h]\footnotesize\center
\begin{tabular}{|c|c|}
\hline
\textbf{Relation Type} & \textbf{Examples}\\
\hline
EMP\_ORG & \texttt{\underline{Indian} \underline{minister}}, \texttt{\underline{employee} of \underline{Microsoft}}\\
PHYS & \texttt{\underline{people} from \underline{the valley}}, \texttt{}\\
PER\_SOC & \texttt{\underline{his} \underline{brother}}, \texttt{\underline{wife} of \underline{the President}}\\
GPE\_AFF & \texttt{\underline{Indian} \underline{singer}}, \texttt{\underline{citizens} of \underline{Japan}}\\
OTHER\_AFF & \texttt{\underline{Christian} \underline{people}}\\
ART & \texttt{\underline{my} \underline{house}}, \texttt{\underline{British} \underline{helicopters}}\\
\hline
\end{tabular}
\label{tab:ACE_relation_examples}
\caption{Examples of ACE 2004 Relation types. The entity mentions involved in the relations, are underlined}
\end{table}

%\section{Introduction}

\subsection{Relation Extraction : Global level Vs Mention level}\label{sec:REtypes}
The term ``Relation Extraction'' is often used in the literature to refer to either global level RE or mention level RE. In this survey, we cover both the types. Global RE system is expected to produce a list of entity pairs for which a certain semantic relation exists. It generally takes a large text corpus as input and produces such a list as output. On the other hand, mention level RE system takes as input an entity pair as well as the sentence which contains it. It then identifies whether a certain relation exists for that entity pair. Consider the entity mentions \texttt{Obama} and \texttt{India} in the sentence : \texttt{\textbf{Obama} is visiting \textbf{India} today.} Here, the mention level RE system would identify the PHYS relation between \texttt{Obama} and \texttt{India}. Consider another sentence : \texttt{\textbf{Obama} likes \textbf{India}'s culture.} Here, mention level RE system should identify that no relation exists between \texttt{Obama} and \texttt{India} in this particular sentence. Automatic context Extraction (ACE) program~\cite{ace2004} called mention level RE with a more appropriate name : Relation Detection and Characterization (RDC).

\subsection{Previous Surveys} 
A comprehensive survey of Information Extraction was presented by Sarawagi~\cite{sarawagi2008information} which covered some RE techniques. % as it is a sub-field of Information Extraction. 
The first dedicated survey of RE techniques was carried out by Bach and Badaskar~\cite{bach2007survey} but it does not cover many of the recent advancements in the field. Another recent survey was presented by de Abreu et al.~\cite{de2013review} which covers various RE techniques used specifically for Portuguese language. Cohen and Hersh~\cite{cohen2005survey} presented a comprehensive survey of early text mining work in biomedical domain, including extraction of biological entities and relations. Zhou et al.~\cite{zhou2014biomedical} surveyed most of the recent biomedical RE approaches.

In this paper, we survey various RE techniques which are classified into several logical categories : (i) supervised techniques including features-based and kernel based methods (Section 2), (ii) a special class of techniques which jointly extract entities and relations (Section 3) (ii) semi-supervised (Section 4), (iii) unsupervised (Section 5), (iv) Open Information Extraction (Section 6) and (v) distant supervision based techniques (Section 7). % that have been used for the task of RE. These techniques can be broadly classified into 3 categories: (a) feature-based; (b) kernel-based; and (c) weakly-supervised. We shall also survey web-based RE. 
%This paper is organized as follows. %In section 2, we discuss a well-known public-domain dataset used for evaluation of RE algorithms. Section 3 discusses two most widely used interpretation of the term Relation Extraction. 
%Supervised RE techniques are discussed in the section 2. A special class of techniques which try to jointly extract entities and relations, are described in the section 3. Semi-supervised and unsupervised RE techniques are presented in the sections 4 and 5, respectively. The recent paradigms of Open Information Extraction and Distant Supervision are discussed in the sections 6 and 7, respectively. %A new paradigm {\em Open Information Extraction} is discussed in the section 6. Distant supervision, a recent approach for reducing supervision efforts, is discussed in the section 7. %In section 10, we present some biomedical domain specific RE techniques. 
Some recent advanced RE techniques are discussed in the section 8. Finally, we conclude in the section 9 by discussing some of the potential future research directions for RE.

\section{Supervised Approaches}\label{sec:supervised}
Supervised approaches focus on RE at the mention level. These approaches require labelled data where each pair of entity mentions is labelled with one of the pre-defined relation types. A special relation type NONE is used to label those pairs where none of the pre-defined relation types hold. 
In general, RE is formulated as a multi-class classification problem with each class corresponding to a different relation type (including NONE).
%In general, supervised approaches learn a multi-class classifier having number of classes equal to number of relation types plus one corresponding to the NONE class. 
These approaches are broadly classified into two types: Feature-based and Kernel-based methods.
\subsection{Feature-based Methods}~\label{sec:featurebased}
In feature-based methods, for each relation instance (i.e. pair of entity mentions) in the labelled data, a set of features is generated and a classifier (or an ensemble of classifiers) is then trained to classify any new relation instance. Kambhatla~\cite{Kam04} described various lexical, syntactic and semantic features for extracting features. Consider the entity pair \texttt{<leaders, Venice>} in the sentence : \texttt{Top leaders of Italy's left-wing government were in Venice.} Table~\ref{tab:features} lists various features derived for this entity pair.

\begin{table}\footnotesize
\begin{tabular}{|p{6.5cm}|p{5cm}|}
\hline
\textbf{Feature Types} & \textbf{Example} \\
\hline
\textbf{Words:} Words of both the mentions and all the words in between & \texttt{M11\_leaders}, \texttt{M21\_Venice}; \texttt{B1\_of}, \texttt{B2\_Italy}, \texttt{B3\_'s}, \texttt{B4\_left-wing}, \texttt{B5\_government}, \texttt{B6\_were}, \texttt{B7\_in}\\
\hline
\textbf{Entity Types:} Entity types of both the mentions & \texttt{E1\_PERSON}, \texttt{E2\_GPE} \\
\hline
\textbf{Mention Level:} Mention types (NAME, NOMINAL or PRONOUN) of both the mentions & \texttt{M1\_NOMINAL}, \texttt{M2\_NAME} \\
\hline
\textbf{Overlap:} \#words separating the two mentions, \#other mentions in between, flags indicating whether the two mentions are in the same NP, VP or PP & \texttt{7\_Words\_Apart}, \texttt{2\_Mentions\_In\_Between} (\texttt{Italy} \& \texttt{government}), \texttt{Not\_Same\_NP}, \texttt{Not\_Same\_VP}, \texttt{Not\_Same\_PP} \\
\hline
\textbf{Dependency:} Words, POS and chunk labels of words on which the mentions are dependent in the dependency tree,  \#links traversed in dependency tree to go from one mentions to another
& \texttt{M1W\_were}, \texttt{M1P\_VBD}, \texttt{M1C\_VP}, \texttt{M2W\_in}, \texttt{M2P\_IN}, \texttt{M2C\_PP}, \texttt{DepLinks\_3} \\
\hline
\textbf{Parse Tree:} Path of non-terminals connecting the two mentions in the parse tree, and the path annotated with head words & \texttt{PERSON-NP-S-VP-PP-GPE}, \texttt{PERSON-NP:leaders-S -VP:were-PP:in-GPE} \\
\hline
\end{tabular}
\caption{Various feature types with examples described by Kambhatla~\cite{Kam04}}
\label{tab:features}
\end{table}

Kambhatla~\cite{Kam04} trained a maximum entropy classifier with 49 classes : two for each relation subtype (ACE 2003 has $24$ relation subtypes and each subtype gives rise to $2$ classes considering order of relation arguments) and a NONE class for the case where the two mentions are not related. Building upon Kambhatla's work, Zhou et al.~\cite{zhou2005} explored some more features to improve the RE performance further. Some of the important additional features are as follows:\\
\noindent\textbf{Word based features:} words between the two mentions (classified into $3$ bins: the first word, last word and other words); first and second words before and after the mentions; headwords of the mentions; flag indicating whether any one of the mention is contained within another\\
\noindent\textbf{Base phrase chunking based features:} phrase heads between the two mentions (classified into $3$ bins: the first, the last and other phrase heads); the first and second phrase heads before and after the mentions; path of phrase labels connecting the two mentions with and without augmentation with head words\\
\noindent\textbf{Features based on semantic resources:} country name list is used to distinguish between {\em citizen\_of} and {\em residence} relation types - when the first (second) mention is a country name, a feature \texttt{CountryET2} (\texttt{ET1Country}) is generated; personal relative trigger word list is used to differentiate $6$ personal social relation subtypes. %({\em Parent}, {\em Grandparent}, {\em Spouse}, {\em Sibling}, {\em Other-Relative} and {\em Other-Personal}). 
It is gathered from WordNet by collecting all the words having the semantic class ``relative''. This list is then classified into different categories representing each of the social relation subtype. The feature \texttt{SC1ET2} (\texttt{ET1SC2}) is generated when the first (second) mention is found in the trigger list where ET2 (ET1) is type of second (first) mention's entity type and SC1 (SC2) is semantic class of the first (second) mention.
%\end{itemize}
%\end{itemize}

Zhou et al.~\cite{zhou2005} employed a SVM classifier using these features and achieved better performance than Kambhatla's system. As SVM is a binary classifier, to achieve multi-class classification {\em one vs. others} strategy is used. They also analysed the results to find contributions by various types of features. The phrase based chunking features were observed to be contributing the most to the increased accuracy. The syntactic features based on dependency tree and parse tree contributed only slightly. A major reason for this is that most of the relations in the ACE data are short-distance relations and simple features like word and chunking features are enough to identify such relations.

A systematic study of the feature space for RE is conducted by Jiang and Zhai~\cite{jiang2007} and they also evaluated the effectiveness of different feature subspaces. They defined a unified graphic representation of the feature space, and experimented with $3$ feature subspaces, corresponding to sequences, syntactic parse trees and dependency parse trees. Experimental results showed that each subspace is effective by itself, with the syntactic parse tree subspace being the most effective. Also, combining the three subspaces did not generate much improvement. They observed that within each feature subspace, using only the basic unit features can already give reasonably good performance and adding more complex features may not improve the performance much.

Some more interesting features are described by Nguyen et al.~\cite{nguyen2007relation} who used SVM for identifying relations between Wikipedia entities. They semi-automatically created list of keywords providing cues for each relation type. E.g. for PHYS relation, words like \texttt{located}, \texttt{headquartered}, \texttt{based} act as keywords. They came up with a novel concept of {\em core tree} to represent any relation instance. This {\em core tree} not only consists of the shortest path connecting two entity mentions in the dependency tree but also additional paths connecting nodes on the shortest path to keywords in the sentence. Then subtrees of this {\em core tree} are mined to act as features.

Chan and Roth~\cite{chan2011exploiting} described an interesting approach for supervised RE based on the observation that all ACE 2004 relation types are expressed in one of several constrained syntactico-semantic structures.\\ %They identified 4 such structures:\\
%\begin{itemize}
%\item 
1. {\em Premodifier}: An adjective or a proper noun modifies another noun (\texttt{Indian minister})\\
%\item 
2. {\em Possessive}: First mention is in possessive case (\texttt{Italy's government})\\
%\item 
3. {\em Preposition}: Two mentions are related via a preposition (\texttt{governor of RBI})\\
%\item 
4. {\em Formulaic}: Two mentions are written is some specific form (\texttt{Mumbai, India})
%\end{itemize}

These structures can be identified by using some simple rules and patterns. %e.g. the structure is {\em Possessive}, if the following patterns are matched:\\
%\begin{itemize}
%\item 
%1. \texttt{M1\_HW 's M2} (e.g. \texttt{RBI's governor})\\
%\item 
%2. \texttt{M1 M2} if POS tag of \texttt{M1} is \texttt{WP\$} or \texttt{PRP\$} (e.g. \texttt{his wife})
%\end{itemize}
%Here, \texttt{M1} and \texttt{M2} are two mentions and \texttt{M1\_HW} is head word of the first mention.
Authors observed that identifying a appropriate syntactico-semantic structure first and then using a specialized RE model which leverages these structures, results in better performance.

One of the major problem that arises in supervised RE methods is that of {\em Class Imbalance}. This happens because number of negative instances (i.e. entity pairs having no meaningful relation) greatly outnumber number of positive instances (i.e. entity pairs having any one of the pre-defined relation type). This {\em Class Imbalance} results in a higher precision and a lower recall as classifiers tend to overproduce the NONE class. Kambhatla~\cite{kambhatla2006minority} presented a novel solution for this problem based on voting among a committee of classifiers that significantly boosts the recall in such situations.

%\textbf{Modelling RE as sequence labelling:} Most of the supervised RE systems employ point classifiers like SVM, MaxEnt. Culotta et al.~\cite{culotta2006integrating} attempted a slightly different RE problem where documents are restricted to be biographical articles and each document has one {\em principal entity} (person whose biography is described in the document) and many other {\em secondary entities} (e.g. parents, spouse, organizations). The goal is for each secondary entity, one has to predict what relation, if any, it has with the principal entity. They treated this problem as a sequence labelling problem (such as POS tagging) where tag for every secondary entity is its relation with the principal entity. Conditional Random Fields (CRF)~\cite{} are used for to obtain these sequence tags. Due to modelling flexibility of CRFs, the feature functions capturing relational patterns from a database of entities can be used in addition to common lexical and syntactic features. Example of such features indicating relational connections between entities can be: {\em father-sibling-son}, {\em father-nationality}. For a secondary entity, these features indicate path of various relations (like {\em father, nationality} etc.) in the database of entities, which connect principal entity to it. Even if such features are not highly precise, authors observed that they still influence the final tagging decision in a positive way.

Once the features are designed, feature-based methods can simply use any classifier from the Machine Learning literature. Most of the efforts in these methods are spent in designing the ``right'' set of features. Arriving at such a features set requires careful analysis of contribution of each feature and knowledge of underlying linguistic phenomena.

\subsection{Kernel Methods}\label{sec:KM}
%Feature-based methods require explicit design of various features and the overall performance largely depends on how effective these features are. 
The overall performance of feature-based methods largely depends on effectiveness of the features designed. The main advantage of kernel based methods is that such explicit feature engineering is avoided. In kernel based methods, kernel functions are designed to compute similarities between representations of two relation instances and SVM (Support Vector Machines) is employed for classification. Various kernel based RE systems propose different representations for relation instances like sequences, syntactic parse trees etc. %differ in terms of how they represent relation instances and how similarity among any two representations is computed. Possible representations are sequences, syntactic parse trees, dependency trees, dependency tree paths, etc. In general, 
Most of the techniques measure the similarity between any two representations (say trees) in terms of number of shared sub-representations (subtrees) between them.

\subsubsection{Sequence Kernel}
Relation instances are represented as sequences and the kernel computes number of shared subsequences between any two sequences. 
Motivated by the string subsequence kernel (Lodhi et al.~\cite{lodhi2002text}), Bunescu and Mooney~\cite{mooney2005} proposed a sequence kernel for RE.
%The string subsequence kernel proposed by Lodhi et al.~\cite{lodhi2002text} was the first subsequence kernel proposed and it has inspired many other sequence kernels. Bunescu and Mooney~\cite{mooney2005} described first such kernel for RE. 
The simplest way to construct a sequence to represent a relation instance is to simply consider the sequence of words from the first mention to the second one in the sentence. Rather than having each sequence element as a singleton word, the authors proposed to generalize each word to a feature vector. Each relation instance is then represented as a sequence of feature vectors, one feature vector for each word. The features come from the following domains:

%\vspace{-2mm}
{\footnotesize
\begin{itemize}
\item $\Sigma_1$: Set of all words  
\item $\Sigma_2$: Set of all POS tags = \{NNP, NN, VBD, VBZ, IN, $\cdots$\}
\item $\Sigma_3$: Set of all generalized POS tags = \{NOUN, VERB, ADJ, ADV, $\cdots$\}
\item $\Sigma_4$: Set of entity types = \{PER, ORG, LOC, GPE, $\cdots$\}
\end{itemize}}
Consider the relation instance formed by the entity pair \texttt{Italy}-\texttt{government} from our example sentence. Table~\ref{tab:seq1} shows the sequence of feature vectors for this instance, where each row is a feature vector.
\begin{table}[h]\footnotesize
\begin{center}
\begin{tabular}{|c|c|c|c|}
\hline
\textbf{Word} & \textbf{POS tag} & \textbf{Generalized POS tag} & \textbf{Entity Type} \\
\hline
\texttt{Italy} & NNP & NOUN & GPE \\
\texttt{'s} & POS & POS & O \\
\texttt{left-wing} & JJ & ADJ & O \\
\texttt{government} & NN & NOUN & ORG \\
\hline
\end{tabular}
\caption{Example of sequence of feature vectors (Sequence $s$)}
\label{tab:seq1}
\end{center}
\end{table}
\vspace{-8mm}
\begin{table}[h]\footnotesize
\begin{center}
\begin{tabular}{|c|c|c|c|}
\hline
\textbf{Word} & \textbf{POS tag} & \textbf{Generalized POS tag} & \textbf{Entity Type} \\
\hline
\texttt{India} & NNP & NOUN & GPE \\
\texttt{'s} & POS & POS & O \\
\texttt{summer} & NN & NOUN & O \\
\texttt{capital} & NN & NOUN & GPE \\
\hline
\end{tabular}
\caption{Example of sequence of feature vectors (Sequence $t$)}
\label{tab:seq2}
\end{center}
\end{table}
\vspace{-4mm}
It is clear that the domain for sequences of feature vectors is $\Sigma_X = \Sigma_1 \times \Sigma_2 \times \Sigma_3.  \times \Sigma_4$. The aim is to design a kernel function which finds shared subsequences $u$ belonging to the domain $\Sigma^*_U=\Sigma_1 \cup \Sigma_2 \cup \Sigma_3 \cup \Sigma_4$. Given two sequences $s, t$ of feature vectors, Bunescu and Mooney~\cite{mooney2005} defined Generalized Subsequence kernel, $K_n(s, t, \lambda)$ which computes number of weighted feature sparse subsequences $u$ of length $n$ such that,
\begin{itemize}\small
\item $u \prec s[ii]$ and $u \prec t[jj]$, for some index sequences $ii, jj$ of length $n$
%\item for some index sequences $ii, jj$ of length $n$
\item the weight of $u$ is $\lambda^{l(ii)+l(jj)}$, where $0<\lambda<1$ and $l(ii)$ is the length of the subsequence which is the difference between largest and smallest index in $ii$. %Sparser the subsequence, more is the penalty it gets.
Sparser the subsequence, lower is its weight.
\end{itemize}
Here, $\prec$ indicates {\em component-wise belongs to} relation, i.e. if $ii=(i_1,i_2,\cdots,i_{|u|})$ and $u \prec s[ii]$ then $u[1] \in s[i_1], u[2] \in s[i_2],\cdots u[|u|] \in s[i_{|u|}]$. Suppose, $ii=(1,2,4)$ then (NNP, \texttt{'s}, NN)$\prec s[ii]$ (Sequence $s$ as shown in Table~\ref{tab:seq1}). Considering the sequences $s$ (Table~\ref{tab:seq1}) and $t$ (Table~\ref{tab:seq2}), some of the shared sparse subsequences of length $n=3$ are:
\begin{center}\footnotesize
(NNP, \texttt{'s}, NN); (NOUN, \texttt{'s}, NN); (NNP, POS, NN); (NOUN, \texttt{'s}, NOUN)%; (NNP, NN, NOUN); and many more
\end{center}
%\begin{itemize}
%\item (NNP, \texttt{'s}, NN)
%\item (NOUN, \texttt{'s}, NN)
%\item (NNP, POS, NN)
%\item (NOUN, \texttt{'s}, NOUN)
%\item (NNP, NN, NOUN)
%\item and many more
%\end{itemize} 
\begin{figure}\center
\includegraphics[scale=0.3]{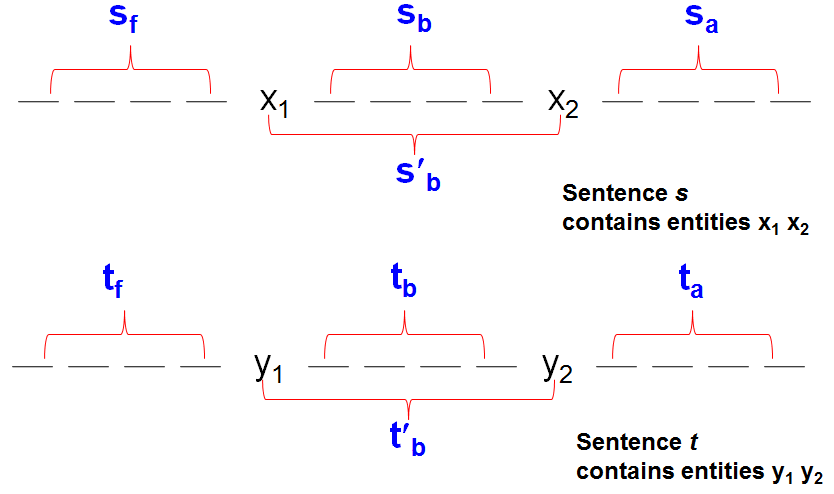}
\label{fig:sequence}
%\caption{Illustration of how sequences are formed for various subkernels of overall relation kernel}
\caption{Illustration of sequence formation for various subkernels of overall relation kernel}
\end{figure}
The generalized subsequence kernel for sequences $s$ and $t$ is computed efficiently using following recursive formulation:
{\footnotesize
\begin{itemize}
\item $K_0^{'}(s,t) = 1$, for all $s,t$
\item $K_i^{"}(sx,ty) = \lambda K_i^{"}(sx,t) + \lambda^2 K_{i-1}^{'}(s,t)\cdot c(x,y)$
\item $K_i^{'}(sx,t) = \lambda K_i^{'}(s,t) + K_i^{"}(sx,t)$
\item $K_n(sx,t) = K_n(s,t) + \sum_j \lambda^2 K_{n-1}^{'}(s,t[1:j-1)\cdot c(x,t[j])$
\end{itemize}}
Here, $x$ and $y$ are feature vectors and $c(x,y)$ is number of common features between $x$ and $y$. $sx$ is a sequence of feature vectors constructed by appending feature vector $x$ to the sequence $s$.

\noindent\textbf{Relation Kernel} ($rK$) is defined as a sum of 4 subkernels, each of which captures a specific type of pattern and is based on the generalized subsequence kernel. The detailed description of these subkernels along with types of patterns they capture are explained with the help of the Figure~\ref{fig:sequence} below:
%\begin{itemize}
%\item 

\noindent\textbf{Fore-Between subkernel ($fbK$)}: Counts number of common fore-between patterns, i.e. number of shared subsequences between $s_fs'_b$ and $t_ft'_b$. %Example of such fore-between pattern is : 
(\texttt{\underline{president} PER \underline{of} ORG})

%\item 
\noindent\textbf{Between subkernel ($bK$)}: Counts number of common between patterns, i.e. number of shared subsequences between $s'_b$ and $t'_b$. %Example of between pattern is : \texttt{PER \textbf{has joined} ORG}
(\texttt{PER \underline{joined} ORG})
%\item 

\noindent\textbf{Between After subkernel ($baK$)} : Counts number of common between-after patterns, i.e. number of shared subsequences between $s'_bs_a$ and $t'_bt_a$. %Example of between pattern is : 
(\texttt{PER \underline{chairman of} ORG \underline{announced}})

%\item 
\noindent\textbf{Modifier subkernel ($mK$)}: When there are no other words in between two entity mentions and the first mention acts as a modifier for the other, modifier patterns are useful. This subkernel counts number of common modifier patterns, i.e. number of shared subsequences between $x_1x_2$ and $y_1y_2$. Example of such modifier pattern is : \texttt{\textbf{Serbian general}}, where the first mention \texttt{\textbf{Serbian}} modifies the second mention \texttt{\textbf{general}}.
%\end{itemize}
The overall relation kernel is defined as:

\vspace{-1mm}
{\footnotesize
\begin{equation*}
rK(s,t) = fbK(s,t) + bK(s,t) + baK(s,t) + mK(s,t)
\end{equation*}}
\vspace{-4mm}

Bunescu and Mooney~\cite{mooney2005} then used SVM classifier based on this relation kernel ($rK$). They also experimented with two different scenarios as follows:\\
\textbf{Scenario 1}: Only one multi-class SVM is trained where each relation type is corresponds to one class and an extra class NONE to represent no-relation cases.\\
\textbf{Scenario 2}: One binary SVM is trained only to decide whether any relation holds or not where all relation types are combined as a single class. Another multi-class SVM is then trained which decides appropriate relation type for positive instances identified by the binary SVM classifier. This scenario was found to be yielding better performance than the other one.

\subsubsection{Syntactic Tree Kernel}
\begin{figure*}
\includegraphics[scale=0.35]{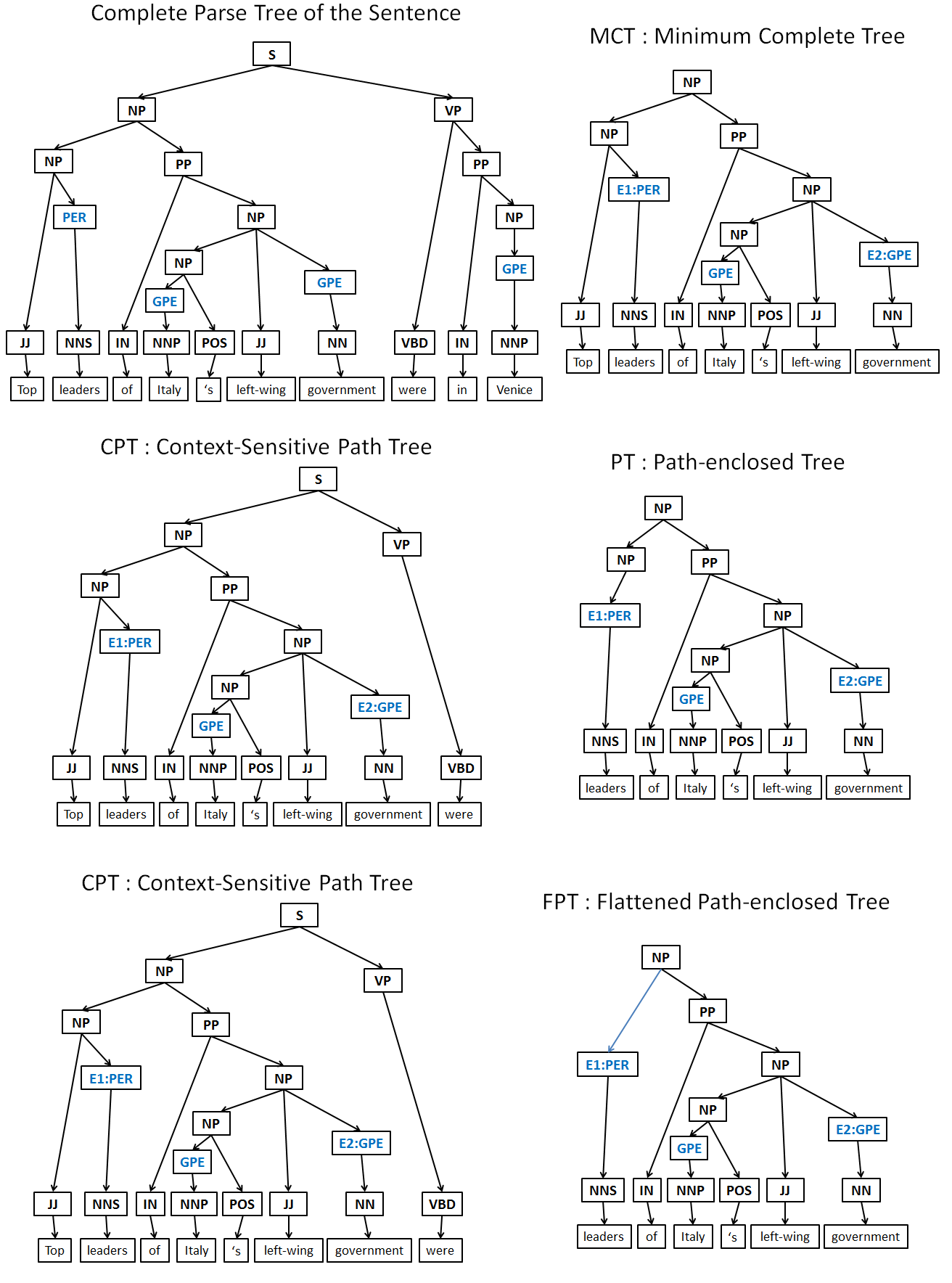}
\caption{Various tree representations described in Zhang et al.~\cite{zhang2006b}}
\label{fig:tree_representations}
\end{figure*}

Structural properties of a sentence are encoded by its constituent parse tree. The tree defines the syntax of the sentence in terms of constituents such as noun phrases (NP), verb phrases (VP), prepositional phrases (PP), POS tags (NN, VB, IN, etc.) as non-terminals and actual words as leaves. Constituent parse tree for our example sentence is shown in the figure~\ref{fig:tree_representations}. The syntax is usually governed by Context Free Grammar (CFG). The task of constructing a constituent parse tree for a given sentence, is called as {\em parsing}. Collins et al.~\cite{collins1998semantic} and Miller et al.~\cite{miller2000novel} proposed statistical parsing models to extract relations from text where they considered the parse trees augmented with information about entities and relations. %We will not cover these techniques in detail, rather 
We focus on the approaches which make use of parse trees already produced by some parsers.

Collins and Duffy~\cite{collins2001convolution} proposed Convolution Parse Tree Kernel $(K_T)$ to compute similarity between any two syntactic trees. It computes number of common subtrees shared by two syntactic parse trees. Here, subtree is defined as any subgraph of a tree which satisfies two conditions - i) it should include more than one node, and ii) entire productions must be included at every node. 
%See the figure~\ref{fig:valid_invalid_subtrees} where valid subtrees are marked with green circles and invalid subtrees are marked with red circles.
%\begin{figure}\center
%\includegraphics[scale=0.35]{valid_invalid_subtrees.png}
%\caption{Examples of valid/invalid subtrees considered by the Convolution Parse Tree Kernel}
%\label{fig:valid_invalid_subtrees}
%\end{figure}
The kernel is designed in such a way that each possible subtree becomes a dimension in the projected space. The image of a syntactic tree $T$ in transformed space is $h(T)=[h_1(T),h_2(T),\cdots h_n(T)]$, where $h_i(T)$ denotes number of occurrences of $i^{th}$ subtree in the tree $T$ and $n$ denotes number of all possible subtrees (subtree vocabulary size). For any two trees $T_1, T_2$, the value of kernel is simply the inner product of their images in the transformed space, i.e. $K_T(T_1, T_2)=h(T_1)\cdot h(T_2)$.

\textbf{Efficient Computation:} It is not feasible to explicitly construct the image vector, as number of all possible subtrees is huge. Hence, the kernel has to be computed efficiently without actually iterating through all possible subtrees. Let $I_i(n) = 1$ if $i^{th}$ subtree is seen rooted at node $n$ and $0$ otherwise. Let $N_1$ and $N_2$ be sets of nodes in trees $T_1$ and $T_2$ respectively.

{\footnotesize
\begin{equation*}
h_i(T_1)=\sum_{n_1 \in N_1}I_i(n_1), h_i(T_2)=\sum_{n_2 \in N_2}I_i(n_2)
\end{equation*}
\begin{equation*}
h(T_1)\cdot h(T_2)=\sum_i h_i(T_1)h_i(T_2)=\sum_{n_1 \in N_1}\sum_{n_2 \in N_2}\sum_i I_i(n_1)h_i(n_2)=\sum_{n_1 \in N_1}\sum_{n_2 \in N_2} C(n_1, n_2)
\end{equation*}}
%\begin{equation*}
%=\sum_{n_1 \in N_1}\sum_{n_2 \in N_2}\sum_i I_i(n_1)h_i(n_2)=\sum_{n_1 \in N_1}\sum_{n_2 \in N_2} C(n_1, n_2)
%\end{equation*}
Here, $C(n_1,n_2)$ counts number of common subtrees rooted at $n_1$ and $n_2$, which can be computed in polynomial time using following recursive definition.\\
%\begin{itemize}
%\item 
1. If the productions at $n_1$ \& $n_2$ are different then, $C(n_1,n_2)=0$\\
%\item 
2. If the productions at $n_1$ \& $n_2$ are same and $n_1,n_2$ are pre-terminals then, $C(n_1,n_2)=1$\\
%\item 
3. If the productions at $n_1$ \& $n_2$ are same and $n_1,n_2$ are not pre-terminals then,

\vspace{-2mm}
{\footnotesize
\begin{equation*}
C(n_1,n_2)=\prod_{j=1}^{nc(n_1)}(1+C(ch(n_1,j),ch(n_2,j))) 
\end{equation*}}
$nc(n)$ denotes number of children of $n$ %(which is same as number of children of $n_2$ as the productions are same) 
and $ch(n,j)$ denotes $j^{th}$ child-node of $n$.
%\end{itemize}

\noindent\textbf{Relation Instance Representation:} A sentence containing $N_e$ entity mentions gives rise to $N_e \choose 2$ relation instances. Hence, it is also important to decide which part of the complete syntactic tree characterizes a particular relation instance. Zhang et al.~\cite{zhang2006a} described five cases to construct a tree representation for a given relation instance which are shown in the figure~\ref{fig:tree_representations}. These representations are for the relation instance constituting entity mentions \texttt{leaders} (E1) and \texttt{government} (E2).\\
1. Minimum Complete Tree (MCT): It is the complete subtree formed by the lowest common ancestor of the two entities.\\
2. Path-enclosed Tree (PT): It is the smallest subtree including both the entities. It can also be described as the subtree enclosed by the shortest path connecting two entities in the parse tree of the sentence.\\
3. Context-sensitive Path Tree (CPT): It is the extended version of PT where one word left of first entity and one word right of second entity are included.\\
4. Flattened Path-enclosed Tree (FPT): It is the modified version of PT where the non-POS non-terminal nodes which are having a single in and out arcs, are bypassed.\\
5. Flattened Context-sensitive Path Tree (FCPT): It is the modified version of CPT where the non-POS non-terminal nodes which are having a single in and out arcs, are bypassed.

In their experimental analysis, Zhang et al.~\cite{zhang2006a} found the Path-enclosed Tree (PT) performs the best when used for computing $K_T$. 
%This work was extended by Zhang et al.~\cite{zhang2006b} by experimenting with another kernel function, the Entity Kernel ($K_E$) which captures similarity of pairs of entity mentions. Each entity mention is characterized by various features such as headword, entity type, entity subtype and mention type (name, nominal and pronoun). Entity Kernel computes number of shared features between two pairs of entity mentions. Using Syntactic Tree Kernel ($K_T$) and Entity Kernel ($K_E$), two composite kernels are constructed to compute similarity between two relation instances $R_1,R_2$:
%\begin{itemize}
%%\item Linear Combination, $K_{LC}(R_1,R_2) = \alpha\cdot NK_E(R_1,R_2) + (1-\alpha) \cdot NK_T(R_1,R_2)$ 
%\item Linear Combination, $K_{LC} = \alpha\cdot NK_E + (1-\alpha) \cdot NK_T$ 
%%\item Polynomial Expansion, $K_{PE}(R_1,R_2) = \alpha \cdot \left( 1 + NK_E(R_1,R_2) \right)^2 + (1-\alpha) \cdot NK_T(R_1,R_2)$
%\item Polynomial Expansion, $K_{PE} = \alpha \cdot \left( 1 + NK_E \right)^2 + (1-\alpha) \cdot NK_T$
%\end{itemize}
%where $NK_E$ ($NK_T$) is the normalized version of Entity (Syntactic Tree) Kernel.
%\begin{equation*}
%NK_E(R_1,R_2)=\frac{K_E(R_1,R_2)}{\sqrt{K_E(R_1,R_1)K_E(R_2,R_2)}}
%\end{equation*}
%Composite Kernels were found to be performing better than the individual kernels and $K_{PE}$ displays the best performance.
Zhou et al.~\cite{zhou122007tree} extended this work further by automatically determining a dynamic context-sensitive tree span for RE by extending the Path-enclosed Tree (PT) to include necessary context information. It also proposed a context-sensitive convolution tree kernel, which in addition to context-free subtrees, considers context-sensitive subtrees also by considering their ancestor node paths as their contexts. Another approach to dynamically determine the tree span, was proposed by Qian et al.~\cite{qian2008exploiting}. They used the information about {\em constituent dependencies} to keep the nodes and their head children along the path connecting the two mentions and removed the other noisy information from the syntactic parse tree. In any Context Free Grammar (CFG) rule, the parent node depends on the head child and this is what the authors called as {\em constituent dependencies}. Another extension to the syntactic tree kernel was proposed by Qian et al.~\cite{qian2007relation} where the parse tree is augmented with entity features such as entity type, subtype, and mention level. Khayyamian et al.~\cite{khayyamian2009syntactic} proposed a generalized version of syntactic tree kernel by Collins and Duffy~\cite{collins2001convolution} for better RE performance. Most of the work in RE using syntactic parse tree kernels is discussed in detail by Zhang et al.~\cite{zhang2008exploring} and Zhou et al.~\cite{zhou2010tree}. Recently, Sun and Han~\cite{sun2014feaure} proposed an extension to the basic syntactic tree kernel, named Feature-enriched Tree Kernel (FTK). Here, the authors proposed to annotate the nodes in a syntactic tree with a set of discriminant features (like WordNet senses, context information, properties of entity mentions, etc.) and FTK is designed to compute the similarity between such enriched trees.

\subsubsection{Dependency Tree Kernel}
\begin{figure}\center
\includegraphics[width=11cm,height=4.5cm]{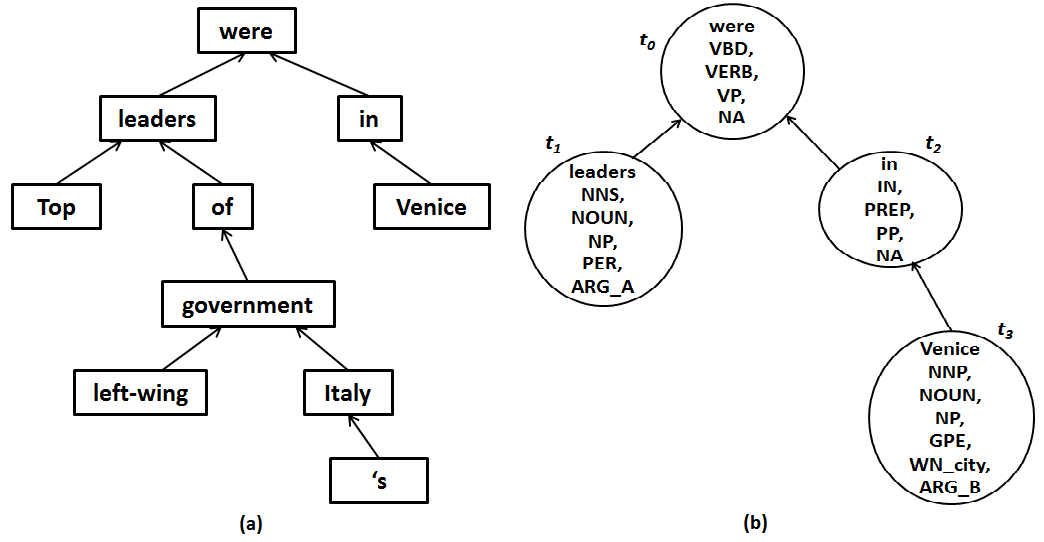}
%\includegraphics[scale=0.35]{aug_dep_tree.png}
%\caption{Augmented dependency tree representation for the relation instance \texttt{leaders-Venice}}
\caption{Dependency Tree for the example sentence and augmented dependency tree representation for the relation instance \texttt{leaders-Venice}}
\label{fig:aug_dep_tree}
\end{figure}

Grammatical relations between words in a sentence are encoded by its dependency tree. Words in the sentence become the nodes in the tree and dependency relations among them become the edges. Each word except the root has exactly one parent in the tree. The directions of the edges are generally shown as pointing from dependent word to its parent. Dependency tree for our example sentence is shown in the figure~\ref{fig:aug_dep_tree}(a). Culotta and Sorensen~\cite{culotta2004} proposed a kernel to compute similarity between two dependency trees. Their work was an extension of the tree kernel proposed by Zelenko et al.~\cite{zelenko2003kernel} which was defined for shallow parse tree representations. 

\noindent\textbf{Relation Instance Representation:} For each entity mention pair in a sentence, smallest subtree of the sentence's dependency tree which contains both the mentions, is considered. Each node in dependency tree is also augmented with additional features such as POS tag, generalized POS tag, chunk tag, entity type, entity level (name, nominal, pronoun), WordNet hypernyms and relation argument (ARG\_A indicating first mention and ARG\_B indicating second mention). In case of our example sentence, the relation instance formed by mention pair \texttt{leaders-Venice} is represented by augmented dependency tree as shown in the figure~\ref{fig:aug_dep_tree}(b).

Formally, a relation instance is represented by a augmented dependency tree $T$ having nodes $\{t_0\cdots t_n\}$ where each node $t_i$ has the features $\phi(t_i)=\{v_1\cdots v_d\}$. Let, $t_i[c]$ denote all children of $t_i$ and $t_i.p$ denote parent of $t_i$. Also, $t_i[jj]$ denotes a particular subset of children of $t_i$ where $jj={j_1, j_2,\cdots j_{l(jj)}}$ (such that $j_1 < j_2 < \cdots < j_{l(jj)}$, $l(jj)$:Length of sequence, Sparseness $d(jj)=j_{l(jj)}-j_1+1$) is an ascending sequence of indices. In case of the tree in the figure~\ref{fig:aug_dep_tree}(b), $t_0[c]=t_0[\{0,1\}]=\{t_1, t_2\}$ and $t_1.p = t_0$. For comparison on any two nodes $t_i,t_j$ two functions are defined,\\
%\begin{itemize}
%\item 
1. \textbf{Matching function ($m(t_i,t_j)$):} It returns $1$ if some important features are shared between $t_i$ and $t_j$, otherwise it returns $0$.\\
%\item 
2. \textbf{Similarity function ($s(t_i,t_j)$):} Unlike the binary matching function, similarity function returns a positive real value as a similarity score between $t_i$ and $t_j$. It is defined as,
{\footnotesize
\begin{equation*}
s(t_i,t_j)=\sum_{v_q \in \phi(t_i)}\sum_{v_r \in \phi(t_j)} C(v_q,v_r)
\end{equation*}}
where $C(v_q,v_r)$ is a compatibility function between two feature values $v_q$ and $v_r$. In the simplest form, $C(v_q,v_r) = 1$ if $v_q=v_r$ and $0$ otherwise.
%\end{itemize}

The overall dependency tree kernel $K(T_1,T_2)$ which measures similarity between two dependency trees $T_1$ and $T_2$ rooted at $t_{10}$ and $t_{20}$ respectively, is defined as follows:\\
$K(T_1,T_2) = 0$, if $m(t_{10},t_{20}) = 0$\\
$K(T_1,T_2) = s(t_{10},t_{20}) + K_c(t_{10}[c],t_{20}[c])$, otherwise\\
where $K_c(t_i[c],t_j[c])$ is a kernel function over children of $t_i$ and $t_j$ which is defined as:
{\footnotesize
\begin{equation*}
\sum_{\substack{ii,jj\\l(ii)=l(jj)}} \lambda^{d(ii)+d(jj)} \left( \sum_{s=1}^{l(ii)}K(t[i_s],t[j_s])\right) \prod_{s=1}^{l(ii)} m(t_i[i_s],t_j[j_s]) 
\end{equation*}}
Intuitively, whenever a pair of {\em matching} nodes is found, all possible {\em matching} subsequences of their children found. Two subsequences are said to be {\em matching} subsequences when all nodes within them are {\em matching} pairwise. Similarity scores of all nodes within such {\em matching} subsequences are then summed up to get overall similarity of children nodes. The constant $0<\lambda<1$ acts as a decay factor that penalizes sparser subsequences. 

A special {\em contiguous} kernel is also defined by the authors which constrains the children subsequences $ii$ such that $d(ii)=l(ii)$. In addition to these sparse ($K_0$) and contiguous ($K_1$) tree kernels, the authors also experimented with a bag-of-words kernel ($K_2$) which treats the tree as a vector of features over nodes, disregarding the tree structure. They also experimented with two composite kernels : $K_3 = K_0+K_1$ and $K_4 = K_1+K_2$, and found that $K_4$ achieves the best performance on the ACE dataset.

Harabagiu et al.~\cite{harabagiu2005shallow} proposed to enhance this dependency tree kernel with the semantic information obtained from shallow semantic parsers using PropBank~\cite{kingsbury2002treebank} and FrameNet~\cite{baker1998berkeley}. There are other approaches which also used kernels defined over dependency trees for RE like the one by Reichartz et al.~\cite{reichartz2009dependency}.

\subsubsection{Dependency Graph Path Kernel}
Bunescu and Mooney~\cite{bunescu2005} proposed a novel dependency path based kernel for RE. The main intuition was that the information required to assert a relationship between two entities in a sentence is typically captured by the shortest path between the two entities in the dependency graph. Kernel is then designed to capture similarity between shortest dependency paths representing two relation instances.
%\begin{figure}\center
%\centering
%%\includegraphics[scale=0.35]{dep_tree_example.png}
%\includegraphics[width=5cm,height=3.5cm]{dep_tree_example.png}
%\caption{Example of Dependency Graph}
%\label{fig:dep_tree}
%\end{figure}
%Consider the sentence : \texttt{Top leaders of Italy's left-wing government were in Venice.} and its corresponding dependency graph (Figure~\ref{fig:aug_dep_tree}(a)). %(Figure~\ref{fig:dep_tree}). 
Consider the dependency graph (figure~\ref{fig:aug_dep_tree}(a)) of our example sentence. 
%Each relation instance is represented by the shortest path between two entities in this graph. 
For the relation instance \texttt{<leaders, Venice>}, the shortest path is :\\ \texttt{leaders}$\rightarrow$\texttt{were}$\leftarrow$\texttt{in}$\leftarrow$\texttt{Venice}. %whereas for the relation instance \texttt{<leaders, Italy>}, the shortest path is \texttt{leaders}$\leftarrow$\texttt{of}$\leftarrow$\texttt{government}$\leftarrow$\texttt{Italy}

Completely lexicalized paths would lead to data sparsity. Hence, words are categorized into word classes with varying degrees of generality. These word classes are POS tags, generalized POS tags, Named Entity types. Following is an example of a generalized path where each word in the sequence has been generalized.

{\footnotesize
\begin{equation}
\begin{bmatrix}
    \texttt{leaders} \\
    NNS \\
    Noun \\
    PER
\end{bmatrix}
\times
\begin{bmatrix}
\rightarrow
\end{bmatrix}
\times
\begin{bmatrix}
    \texttt{were} \\
    VBD \\
    Verb 
\end{bmatrix}
\times
\begin{bmatrix}
\leftarrow
\end{bmatrix}
\times
\begin{bmatrix}
    \texttt{in} \\
    IN 
\end{bmatrix}
\times
\begin{bmatrix}
\leftarrow
\end{bmatrix}
\times
\begin{bmatrix}
    \texttt{Venice} \\
    NNP \\
    Noun \\
    GPE
\end{bmatrix}
\label{eq:gp1}
\end{equation}}

Each possible path is considered a feature. The above generalized path gives rise to features such as \texttt{leaders}$\rightarrow$\texttt{were}$\leftarrow$\texttt{in}$\leftarrow$\texttt{Venice}, \texttt{NNS}$\rightarrow$\texttt{were}$\leftarrow$\texttt{in}$\leftarrow$\texttt{Venice}, \texttt{NNS}$\rightarrow$\texttt{VBD}$\leftarrow$\texttt{in}$\leftarrow$\texttt{GPE} and so on. There are $4\times1\times3\times1\times2\times1\times4 = 96$ such features. Shorted Dependency Path Kernel computes number of common path features shared by two relation instances. Let two relation instances $R_1$ and $R_2$ be represented by their respective shortest paths: $R_1 = x_{11}x_{12}\cdots x_{1m}$, $R_2 = x_{21}x_{22}\cdots x_{2n}$, then the kernel is computed as,\\
$K(R_1, R_2) = \prod_{i=1}^{n} c(x_{1i},x_{2i})$ when $m = n$\\
$K(R_1, R_2) = 0$ when $m \neq n$

Consider another generalized path as follows:
{\footnotesize
\begin{equation}
\begin{bmatrix}
    \texttt{John} \\
    NNP \\
    Noun \\
    PER
\end{bmatrix}
\times
\begin{bmatrix}
\rightarrow
\end{bmatrix}
\times
\begin{bmatrix}
    \texttt{went} \\
    VBD \\
    Verb 
\end{bmatrix}
\times
\begin{bmatrix}
\leftarrow
\end{bmatrix}
\times
\begin{bmatrix}
    \texttt{to} \\
    IN 
\end{bmatrix}
\times
\begin{bmatrix}
\leftarrow
\end{bmatrix}
\times
\begin{bmatrix}
    \texttt{London} \\
    NNP \\
    Noun \\
    GPE
\end{bmatrix}
\label{eq:gp2}
\end{equation}}
The value of Dependency Path Kernel for the above two relation instances (Eq.~\ref{eq:gp1}, Eq.~\ref{eq:gp2}) is computed to be $2\times1\times2\times1\times1\times1\times2=8$, i.e. number of common featured shared by both the relation instances is $8$. The dependency path kernel explained above imposes a hard constraint that the two paths should have exactly the same number of nodes. In order to make it more flexible, Wang~\cite{wang2008re} proposed a {\em convolution dependency path kernel} which finds number of common subsequences shared by two dependency path sequences. 

\subsubsection{Composite Kernels}
%Individual kernels (like dependency tree kernel, syntactic tree kernel, etc.) can be combined together to form composite kernels. This is useful because 
A composite kernel combines the information captured by the individual kernels. %(like dependency tree kernel, syntactic tree kernel. 
For example, a composite kernel which combines syntactic tree kernel and sequence kernel, uses syntactic information captured by the tree kernel as well as lexical information captured by the sequence kernel. It is important to ensure that the combination of two individual kernels is also a valid kernel function. Some of the valid ways of combining two individual kernels are : sum, product, linear combination.

Zhang et al.~\cite{zhang2006a} used syntactic tree kernel for RE. This work was extended by Zhang et al.~\cite{zhang2006b} by experimenting with another kernel function, the Entity Kernel ($K_E$) which captures similarity of pairs of entity mentions. Each entity mention is characterized by various features such as headword, entity type, entity subtype and mention type (name, nominal and pronoun). Entity Kernel computes number of shared features between two pairs of entity mentions. Using Syntactic Tree Kernel ($K_T$) and Entity Kernel ($K_E$), two composite kernels are constructed to compute similarity between two relation instances $R_1,R_2$:
{\small
\begin{itemize}
%\item Linear Combination, $K_{LC}(R_1,R_2) = \alpha\cdot NK_E(R_1,R_2) + (1-\alpha) \cdot NK_T(R_1,R_2)$ 
\item Linear Combination, $K_{LC} = \alpha\cdot NK_E + (1-\alpha) \cdot NK_T$ 
%\item Polynomial Expansion, $K_{PE}(R_1,R_2) = \alpha \cdot \left( 1 + NK_E(R_1,R_2) \right)^2 + (1-\alpha) \cdot NK_T(R_1,R_2)$
\item Polynomial Expansion, $K_{PE} = \alpha \cdot \left( 1 + NK_E \right)^2 + (1-\alpha) \cdot NK_T$
\end{itemize}
where $NK_E$ ($NK_T$) is the normalized version of Entity (Syntactic Tree) Kernel.
\begin{equation*}
NK_E(R_1,R_2)=\frac{K_E(R_1,R_2)}{\sqrt{K_E(R_1,R_1)K_E(R_2,R_2)}}
\end{equation*}
}
Composite Kernels were found to be performing better than the individual kernels and $K_{PE}$ displays the best performance. Normalization of individual kernels before combining is necessary to ensure that value of one kernel does not overwhelm the value of another.

Zhao and Grishman~\cite{zhao2005extracting} presented a RE approach which combines information from three different levels of NLP processing: tokenization, sentence parsing and deep dependency analysis. Individual kernel functions are designed to capture each source of information. Then composite kernels are developed to combine these individual kernels so that processing errors occurring at one level can be overcome by information from other levels. Nguyen et al.~\cite{nguyen2009convolution} investigated effectiveness of combining various kernels capturing syntactic and semantic information. Syntactic information was captured by individual kernels based on constituent and dependency trees, whereas semantic information is captured by entity types and lexical sequences. Wang et al.~\cite{wang2011relation} proposed a new composite kernel for RE which uses a sub-kernel defined using {\em relation topics}. A training set of around 7000 existing Wikipedia relations is automatically created (using a technique similar distant supervision explained later) by making use of Wikipedia infoboxes. Then, the {\em relation topics} are defined over these existing relations. By leveraging this knowledge extracted from the Wikipedia relation repository, the authors reported a significant improvement in RE performance. Wang et al.~\cite{wang2012relation} further explained the application of this kernel based on {\em relation topics} in the Question Answering framework {\em DeepQA}.

%\begin{table}\footnotesize\center
%\begin{tabular}{|c|c|c|c|c|}
%\hline
%%Type & Approach & Precision & Recall & F-measure \\
%Type & Approach & P & R & F \\
%\hline
%\multirow{2}{*}{Features-based} & Jiang and Zhai~\cite{jiang2007} & 0.737 & 0.694 & 0.715 \\
% & Chan and Roth~\cite{chan2011exploiting} & 0.754 & 0.68 & 0.715 \\
%\hline
%\multirow{3}{*}{Kernel-based} & Zhao and Grishman~\cite{zhao2005extracting} & 0.692 & 0.705 & 0.7035 \\ 
% & Zhang et al.~\cite{zhang2006b} & 0.761 & 0.684 & 0.721 \\
% & Zhou et al.~\cite{zhou122007tree} & 0.822 & 0.702 & 0.758 \\
% & Nguyen et al.~\cite{nguyen2009convolution} & 0.766 & 0.67 & 0.715 \\
% & Qian et al.~\cite{qian2007relation} & 0.792 & 0.674 & 0.728 \\
% & Qian et al.~\cite{qian2008exploiting} & 0.83 & 0.72 & \textbf{0.771} \\
%\hline
%\end{tabular}
%\caption{Comparative RE performance of various supervised approaches on 7 major relation types in ACE 2004 dataset (5-fold cross validation)}
%\label{tab:RE_comparison}
%\end{table}

\begin{table}\footnotesize\center
\begin{tabular}{|c|p{7cm}|c|c|c|}
\hline
%Type & Approach & Precision & Recall & F-measure \\
Type & Approach & P & R & F \\
\hline
Features & LX, ST and DT based features~\cite{jiang2007} & 0.737 & 0.694 & 0.715 \\
\cline{2-5}
based & Additional features based on Syntactico-Semantic structures~\cite{chan2011exploiting} & 0.754 & 0.68 & 0.715 \\
\hline
 & Composite kernel combining individual LX, ST and DT kernel ~\cite{zhao2005extracting} & 0.692 & 0.705 & 0.7035 \\ 
\cline{2-5}
Kernel & Composite kernel combining ST and EN kernels~\cite{zhang2006b} & 0.761 & 0.684 & 0.721 \\
\cline{2-5}
based & ST kernel with dynamically determined tree span~\cite{zhou122007tree} & 0.822 & 0.702 & 0.758 \\
\cline{2-5}
 & Composite kernel combining ST and DT kernels along with semantic information~\cite{nguyen2009convolution} & 0.766 & 0.67 & 0.715 \\
\cline{2-5}
 & ST kernel where parse tree is augmented with entity features~\cite{qian2007relation} & 0.792 & 0.674 & 0.728 \\
\cline{2-5}
 & ST kernel with dynamically determined tree span~\cite{qian2008exploiting} & 0.83 & 0.72 & \textbf{0.771} \\
\hline
\end{tabular}
\caption{Comparative RE performance of various supervised approaches on 7 major relation types in ACE 2004 dataset (5-fold cross validation, LX: lexical, ST: syntactic tree, DT: dependency tree)}
\label{tab:RE_comparison}
\end{table}
\subsection{Evaluation}\label{sec:Eval}
Some of the papers in supervised RE report their performance on non-standard datasets, but some others report their performance on standard datasets such as ACE 2003 and ACE 2004. Similar to multi-class classification, the performance of supervised RE systems is evaluated in terms of {\em precision}, {\em recall} and {\em F-measure} of non-NONE classes. The table~\ref{tab:RE_comparison} shows performance of various approaches on the ACE 2004 dataset with 5-fold cross-validation. Even though the same dataset is used by various approaches, the actual splits/folds used in the 5-fold cross-validation might be different. Nevertheless, these figures provide a rough idea of comparative performances of these approaches. It can be observed that the kernel-based methods generally outperform the feature-based methods and among the various kernel-based methods, the methods based on syntactic tree kernel perform the best.

\section{Joint Extraction of Entities and Relations}\label{sec:JL}
All of the RE techniques explained in the previous section assume that the knowledge about boundaries and types of entity mentions are known before hand. If such knowledge is not available, in order to use these techniques for extracting relations, one needs to first apply some entity mentions extraction technique. Once entity mentions and their entity types are identified, then RE techniques can be applied. Such a ``pipeline'' method is prone to propagation of errors from the first phase (extracting entity mentions) to the second phase (extracting relations). To avoid this propagation of errors, there is a line of research which models or extracts entities and relations jointly.

\subsection{Integer Linear Programming based Approach}\label{sec:ILP}
Roth and Yih~\cite{roth2004linear} proposed a model, which first learns independent local classifiers for entity extraction and RE. During inference, given a sentence, a global decision is produced such that the domain-specific or task-specific constraints are satisfied. A simple example of such constraints is : both the arguments of the \texttt{PER-SOC} relation should be \texttt{PER}. Consider the sentence - \texttt{John married Paris.} Here, the entity extractor identifies two mentions \texttt{John} and \texttt{Paris} and also predicts entity types for these mentions. For the first entity, let the predicted probabilities be : Pr(\texttt{PER}) = 0.99 and Pr(\texttt{ORG}) = 0.01. For the second entity, let the predicted probabilities be : Pr(\texttt{GPE}) = 0.75 and Pr(PER) = 0.25. Also, the relation extractor identifies the relation \texttt{PER-SOC} between the two mentions. If we accept the best suggestions given by the local classifiers, then the global prediction is that the relation \texttt{PER-SOC} exists between the \texttt{PER} mention \texttt{John} and the \texttt{GPE} mention \texttt{Paris}. But this violates the domain-constraint mentioned earlier. Hence the global decision which satisfies all the specified constraints would be to label both the mentions as \texttt{PER} and mark the \texttt{PER-SOC} relation between them. This problem of taking a global decision consistent with the constraints, is solved by using a Integer Linear Programming approach by Roth and Yih~\cite{roth2004linear}. This Integer Linear Program minimizes the sum of assignment cost function and constraint cost function. The assignment cost function is designed in such a way that if the most probable prediction of a local classifier is chosen then the least cost is incurred. The constraint cost function is designed to impose cost for breaking constraints between connected entities and relations.

The experiments reported significant improvement in the RE performance using ILP for global inference, as compared to the {\em pipeline} method which first identifies entity types and using these predicted types relation classifier is run. This global inference approach even improves quality of entity classification which is impossible in the {\em pipeline} approach. Roth and Yih~\cite{roth2007global} described the global inferencing for entity and relation extraction in further details where they explore some other global inferencing techniques than ILP like the Viterbi Algorithm. Chan and Roth~\cite{chan2010exploiting} proposed an extension to the original ILP framework for incorporating background knowledge such as hierarchy of relation types, co-reference information, etc. The ILP based approach for global inferencing was also used by Choi et al.~\cite{choi2006joint} for joint extraction of entities and relations in the context of opinion information extraction. Here, entities were of two types - source of the opinion and expression of opinion. The only relation considered was the linking relation between the two entities.

\subsection{Graphical Models based Approach}\label{sec:GM}
The first attempt of using graphical models approach for jointly identifying entities and relations was by Roth and Yih~\cite{roth2002probabilistic}. They proposed a framework where {\em local} independent classifiers are learned for entities and relations identification. The dependencies between entities and relations are encoded though a bayesian belief network which is a bipartite, directed acyclic graph. Entities are represented as nodes in one layer in bipartite graph whereas relations are represented as nodes in the other layer. %Using the same notations as described in the ILP approach, 
Each relation instance node $R_{ij}$ has two incoming edges from its argument entity instance nodes $E_i, E_j$. %The figure~\ref{fig:BN} shows such a bayesian network for a sentence having two entities. 
%Here, $X$ represents various features characterizing the sentence. 
Given the feature vector $X$ which characterizes the sentence, the local entity and relation classifiers are used to compute $Pr(E_i|X)$ and $Pr(R_{ij}|X)$, respectively. The constraints are encoded through the conditional probabilities $Pr(R_{ij}|E_i,E_j)$, which can be set manually or estimated from the entities and relations labelled corpus.
%\begin{center}
%\begin{figure}\center
%\includegraphics[scale=0.3]{BN.png}
%\label{fig:BN}
%\caption{Bayesian Network representing entities and relations in a sentence having 2 nodes}
%\end{figure}
%\end{center}
The joint probability of the nodes in the bayesian network is maximized to get the most probable label assignments for entity and relation nodes, i.e. $(e_1,e_2,\cdots e_n,r_{12},r_{21},\cdots r_{n(n-1)})$ would be,

{\footnotesize
\begin{equation}
arg\max_{e_i,r_{jk}}Pr(E_1,E_2,\cdots E_n,R_{12},R_{21},\cdots R_{n(n-1)})
\end{equation}}
The joint probability expression involving $Pr(E_i|X)$, $Pr(R_{ij}|X)$ and $Pr(R_{ij}|E_i,E_j)$ is not very clear as it is not explicitly mentioned by the authors. They experimented with two specific relations {\em Born\_in} and {\em Kill} and found that performance of relation classification using bayesian network is better than the independent relation classifier. But similar improvement for entity classification using was not observed. Possible reason for this can be the fact that very few entities were involved in some relation in the datasets used by the authors.

%The joint probability is:
%\begin{equation}
%Pr(E_1,E_2,\cdots E_n,R_{12},R_{21},\cdots R_{n(n-1)}|X)=\left(\prod_{\substack{i,j\\i\neq j}}Pr(R_{ij}|E_i,E_j)\right)
%\end{equation}

Yu and Lam~\cite{yu2010jointly} proposed a framework based on undirected, discriminative probabilistic graphical model to jointly perform the tasks of entity identification and relation extraction. Moreover, unlike most of the other approaches, the knowledge about entity mention boundaries is not assumed and is incorporated as a part of the model. Only thing that simplifies the problem a little is that the relations are always assumed to be between a principal entity and other secondary entities in the sentence, as the focus is on the encyclopaedia articles where each article is about a principle entity. i.e. arbitrary relations among the secondary entities are not allowed. %They estimated the model parameters by maximizing the regularized log likelihood of the data. Also, they proposed a new inference method (collective iterative classification), to find the most likely assignments for both entities and relations.

%All the approaches for joint modelling described till now, only focus on two tasks at a time, i.e. entity extraction and relation extraction. 
Most of the approaches for joint modelling only focus on two tasks at a time, i.e. entity extraction and relation extraction. Singh et al.~\cite{singh2013joint} is the first approach which even models co-references jointly with entity mentions and relations. The task of co-reference resolution is to link entity mentions within a document which refer to the same real-word entity. They proposed a single, joint undirected graphical model that represents the various dependencies between these three tasks. Unlike most other approaches for RE where the modelling is at a sentence level, in this approach a model captures all the entity mentions within a {\em document} along with the relations and co-references amongst them. The challenge here is to handle such a large number of variables in a single model, which is addressed by the authors through an extension to belief propagation algorithm that sparsifies the domains of variables during inference.

\subsection{Card-Pyramid Parsing}
\begin{figure}\center
\includegraphics[scale=0.3]{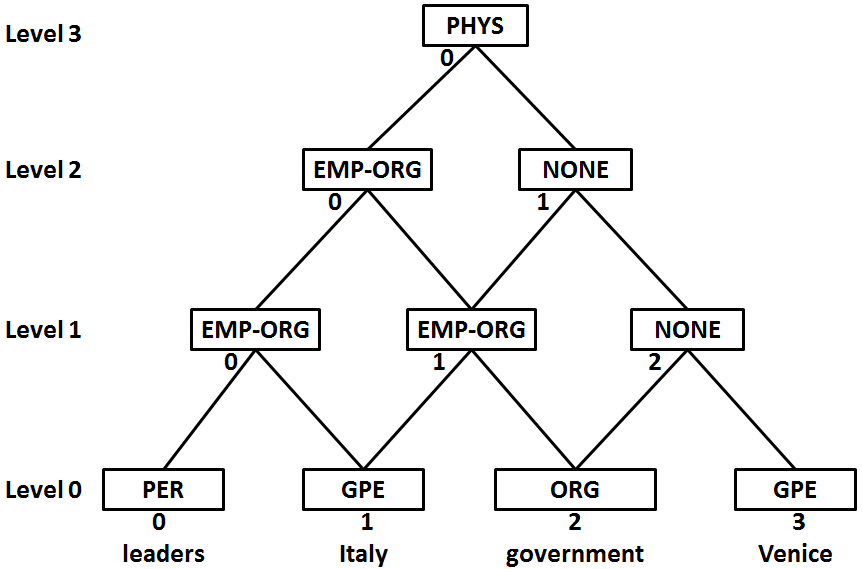}
\caption{Card-pyramid graph for our example sentence}
\label{fig:card_pyramid}
\end{figure}

Another interesting approach for joint extraction of entities and relations was proposed by Kate and Mooney~\cite{kate2010joint} and used a graph (not probabilistic graphical model) called as {\em card-pyramid}. The graph is so called because it encodes mutual dependencies among the entities and relations in a graph structure which resembles pyramid constructed using playing cards. This is a tree-like graph which has one root at the highest level, internal nodes at intermediate levels and leaves at the lowest level. Each entity in the sentence correspond to one leaf and if there are $n$ such leaves then the graph has $n$ levels. Each level $l$ contains one less node than the number of nodes in the $(l-1)$ level. The node at position $i$ in level $l$ is parent of nodes at positions $i$ and $(i+1)$ in the level $(l-1)$. Each node in the higher layers (i.e. layers except the lowest layer), corresponds to a possible relation between the leftmost and rightmost nodes under it in the lowest layer. Figure~\ref{fig:card_pyramid} shows this {\em card-pyramid} graph for our example sentence. 

The aim is to jointly label the nodes in the card-pyramid graph. The authors propose a parsing algorithm analogous to the bottom-up CYK parsing algorithm for Context Free Grammar (CFG) parsing. The grammar required for this new parsing algorithm is called Card-pyramid grammar and its consists of following production types:\\
%\begin{itemize}
%\item 
1. \textbf{Entity Productions} of the form $EntityType\rightarrow Entity$, e.g. \texttt{PER}$\rightarrow$\texttt{leaders}. A local entity classifier is trained to compute the probability that entity in the RHS being of the type given in the LHS of the production.\\
%\item 
2. \textbf{Relation Productions} of the form $RelationType\rightarrow EntityType1$ $EntityType2$, e.g. \texttt{PHYS}$\rightarrow$\texttt{PER} \texttt{GPE}. A local relation classifier is trained to predict the probability that the relation type in the LHS holds between the two entities in the RHS of the production.
%\end{itemize}

Given the entities in a sentence, the card-pyramid grammar and the local entity and relation classifiers, the card-pyramid parsing algorithm attempts to find the most probable labelling of all of its nodes which corresponds the entity and relation types.

\subsection{Structured Prediction}
%In most of the approaches for joint extraction of entities and relations described till now, it was assumed that the boundaries of the entity mentions were known. 
In most of the approaches for joint extraction of entities and relations, it is assumed that the boundaries of the entity mentions are known. Li and Ji~\cite{li2014incremental} presented an incremental joint framework for simultaneous extraction of entity mentions and relations, which also incorporates the problem of boundary detection for entity mentions. Earlier approaches modelled independent {\em local} classifiers for identifying entities and relations. Even though optimal global decision taken later, interaction between entity extraction and RE modules is prohibited during training. Hence, the authors proposed to re-formulate this problem as a structured prediction problem. They try to predict the output structure ($y \in Y$) for a given sentence ($x\in X$), where this structure can be viewed as a graph modelling entity mentions as nodes and relations as directed arcs with relation types as labels. Following linear model is used to predict the most probable structure $y'$ for $x$ where $f(x,y)$ is the feature vector that characterizes the entire structure.

{\footnotesize
\begin{equation*}
y' = arg\max_{y \in Y(x)} f(x,y)\cdot \vec{w}
\end{equation*}}
The score of each candidate assignment is defined as the inner product of the feature vector $f(x,y)$ and feature weights $\vec{w}$. The number of all possible structures for any given sentence can be very large and there does not exist a polynomial-time algorithm to find the best structure. Hence, they apply beam-search to expand partial configurations for the input sentence incrementally to find the structure with the highest score. For decoding, they employed the idea of semi-Markov chain proposed by Sarawagi and Cohen~\cite{Sarawagi:2004}, in which each state corresponds to a segment of the input sequence, instead of treating individual tokens/words as states. %Also, instead of exact inference, they opted for beam search for incremental decoding. 
%They maintain $k$-best partial assignments ending at the $i^{th}$ token, denoted as $B[i]$. These can be selected as top $k$ partial assignments $y'$ according to the score $f(x,y')\cdot \vec{w}$, where $y'$ is from the set:
%{\small
%\begin{equation*}
%\{y_{[1:i]} | y_{[1:i-d]} \in B[i-d], d = 1\cdots d' \}
%\end{equation*}}
%Here, $d'$ is upper bound on entity mention length. To estimate the feature weights, structured perceptron~\cite{Collins:2002} was used. It is an extension of the standard perceptron for structured prediction, as the learning framework. The learning algorithm is as described in the Table~\ref{tab:struct_perceptron_algo}.
%\begin{table}[h]\footnotesize
%\centering 
%\begin{tabular}{p{\textwidth}}
%\hline
%\textbf{Input:} Training set $D = \{ (x, y)^i \}$ for $i=1\cdots N$, Maximum no. of iterations $T$\\
%\textbf{Output:} Model parameters $\vec{w}$\\
%%\begin{itemize}
%%\item 
%1. $\vec{w} \leftarrow 0$\\
%%\item 
%2. \textbf{for} $t\leftarrow 1\cdots T$ \textbf{do}\\
%	%\begin{itemize}
%	%\item 
%	\hspace{3mm}2.1 \textbf{foreach} $(x,y) \in D$ \textbf{do}\\
%		%\begin{itemize}		
%			%\item 
%			\hspace{6mm}2.1.1 $(x,y',z) \leftarrow BeamSearch(x,y,\vec{w})$ (Here, $y'$ is the prefix of the gold-standard which falls out of the beam and $z$ is the top assignment in the beam)\\
%			%\item 
%			\hspace{6mm}2.1.2 \textbf{if} $z \neq y$ \textbf{then} $w \leftarrow w + f(x,y') - f(x,z)$\\
%		%\end{itemize}
%	%\end{itemize}
%%\end{itemize}
%\tabularnewline \hline
%\end{tabular}\caption{Structured Perceptron Training Algorithm.}
%\label{tab:struct_perceptron_algo}
%\end{table}

\noindent\textbf{Features:} Along with various {\em local} features used for entity and relation extraction, a major advantage of this framework is that arbitrary features for both the tasks can be easily exploited. Some of {\em global} features used for entity extraction try to capture long distance dependencies among the entity mentions like,\\
%\begin{itemize}
%\item 
1. Co-reference consistency: Co-reference links between two segments are determined in the same sentence using some simple heuristic rules. A global feature is encoded to check whether two co-referential segments share the same entity type.\\
%\item 
2. Neighbour coherence: The entity types of the two neighbouring segments are linked together as a global feature.\\
%\item 
3. Part-of-whole consistency:  If an entity mention is semantically part of another mention (connected by a {\em prep\_of} dependency link), they should be assigned the same entity type. e.g., in \texttt{some of Italy's leaders}, \texttt{some} and \texttt{leaders} should get same entity type \texttt{PER}.\\
%\end{itemize}
Some of the {\em global} features designed for RE are:\\
%\begin{itemize}
%\item 
1. Triangle constraint: Multiple entity mentions are unlikely to be fully connected with the same relation type. A negative feature was used to penalize any structure that contains this type of formation.\\
%\item 
2. Inter-dependent compatibility: If two entity mentions are connected by a dependency link, they tend to have compatible relations with other entities. e.g., in the sentence \texttt{John and Mary visited Germany}, the {\em conj\_and} dependency link between the mentions \texttt{John} and \texttt{Mary} indicates that they may share the same relation type with some third entity mention \texttt{Germany}.
%\end{itemize}

Another approach for joint extraction of entities and relations was proposed by Miwa and Sasaki~\cite{miwa2014modeling} which uses a table structure. This table represents entity and relation structures in a sentence. If the number of words in a sentence is $n$, then the table is $n\times n$ lower triangular matrix where the $i^{th}$ diagonal cell represents the entity type of the $i^{th}$ word. Any $(i,j)$ cell represents the relation type (if any) between the entity mention headed at $i^{th}$ word and another entity mention headed at $j^{th}$ word. With this table representation, the joint extraction problem is mapped to a table-filling problem where labels are assigned to the cells in the table. Similar to the approach of Li and Ji~\cite{li2014incremental}, various local and global features are captured to assign a score to any labels assignment in a table.

Majority of the approaches which jointly extract entities and relations, report a significant improvement over the basic pipeline approach. Joint extraction not only improves performance of relation extraction but also proves to be effective for entity extraction. Because, unlike pipeline methods, joint model facilitates the use of relations information for entity extraction. It is difficult to compare various methods for joint modelling because no single, standard dataset is used for reporting results. Some of these approaches (like~\cite{roth2002probabilistic,roth2004linear,kate2010joint}) achieve joint modelling only through joint inference as local classifiers for entities and relations are trained independently. Some recent approaches (like~\cite{li2014incremental,singh2013joint}) perform actual joint learning where a single model is learned for extracting both entities and relations. There are a few but consistent contributions to this line of research over the time and still there is a scope for more sophisticated joint models in future.

\section{Semi-supervised Approaches}
%Supervised approaches for RE require labelled data containing gold-standard entities and relation labels. 
It is cost, effort and time intensive task to generate labelled data for RE. Major motivation behind designing semi-supervised techniques is two-fold: i) to reduce the manual efforts required to create labelled data; and ii) exploit the unlabelled data which is generally easily available without investing much efforts. In this section, we describe some major semi-supervised RE approaches.
\subsection{Bootstrapping Approaches}
Generally, bootstrapping algorithms require a large unlabelled corpus and a few {\em seed} instances of the relation type of interest. e.g. in order to learn model/patterns for extracting the relation {\em CaptialOf}, the seed examples can be \texttt{<Beijing, China>}, \texttt{<New Delhi, India>}, \texttt{<London, England>} etc. Given these seed examples, a bootstrapping algorithm is expected to extract similar other entity pairs having the same relation type, e.g. \texttt{<Paris, France>}. The first such bootstrapping algorithm named DIPRE (Dual Iterative Pattern Relation Expansion) was proposed by Brin~\cite{brin1999}. The intuition behind this algorithm is {\em Pattern Relation Duality}, which is:
\begin{itemize}\small
\item Given a good set of patterns, a good set of tuples (entity pairs following a certain relation type) can be found.
\item Given a good set of tuples, a good set of patterns can be learned
\end{itemize}
Combination of above two properties provides great power and it is the basis of the DIPRE algorithm. Table~\ref{tab:DIPRE} shows an overview of DIPRE algorithm.

\begin{table}[h]\footnotesize
\begin{tabular}{p{\textwidth}}
\hline
\textbf{Input:} Seed set $S$ of tuples, i.e. entity pairs known to be related with certain relation type $R$\\
\textbf{Output:} Set $S$ grown over multiple iterations\\
1. Find all {\em occurrences} of the tuples from the seed set $S$ on the Web\\
2. Learn {\em patterns} from these {\em occurrences}\\
3. Search the web using these {\em patterns} and find new tuples and add to the set $S$\\
4. Go to step 1 and iterate till there are no new tuples to be added\\
\hline
\end{tabular}
\caption{Overview of DIPRE~\cite{brin1999} algorithm}
\label{tab:DIPRE}
\end{table}
{\em Patterns} for capturing relation type $R$ between two entities $E_1$ and $E_2$ are represented as a 5-tuple: {\small {\em (order, urlprefix, prefix, middle, suffix)}}. Here, {\em order} is a boolean value and other values are strings. When {\em order} = true, a pair of entities $(E_1, E_2)$ matches the above pattern if URL of the page matches {\em urlprefix} and contains the text : {\small $<${\em prefix}$>E_1<${\em middle}$>E_2<${\em suffix}$>$}
%When {\em order} = false, a pair of entities $(E_1, E_2)$ matches the above pattern if URL of the page matches {\em urlprefix} and contains the text as follows:

%{\footnotesize \begin{center}
%$<${\em prefix}$>E_2<${\em middle}$>E_1<${\em suffix}$>$
%\end{center}}
Example of such a pattern is {\em (true, ``en.wikipedia.org/wiki/'', \texttt{City of}, \texttt{is capital of}, \texttt{state})} and it matches a text like \texttt{City of \textbf{Mumbai} is capital of \textbf{Maharashtra} state.}

Starting with just $3$ seed examples of {\em author, book} pairs and using the corpus of around $24$ million web pages, the DIPRE algorithm was able to generate a list of $15257$ unique {\em author, book} pairs.

Agichtein and Gravano~\cite{agichtein2000snowball} built upon the bootstrapping based idea of DIPRE and they developed a system named {\em SnowBall}. The two main aspects on which {\em SnowBall} proposed advancements over DIPRE are : i) Pattern representation and generation; and ii) Evaluation of patterns and tuples.

\noindent\textbf{Pattern Representation and Generation:} One of the key advancement that {\em SnowBall} proposed over the DIPRE, was the inclusion of named entity tags (PER, ORG, LOC, etc.) in the patterns. The presence of named entities in the patterns made them more meaningful and reduced the number of false positives extracted by them.%, e.g. consider the pattern \texttt{<PER> from <LOC>}. This pattern will not match any string followed by \texttt{from}, but it will match only those strings followed by \texttt{from} which are identified as \texttt{LOC} by the NER tagger. Hence, the text fragment \texttt{John from India} will be matched but \texttt{John from IBM} won't be matched because \texttt{India} is identified by NER as \texttt{LOC} whereas \texttt{IBM} is identified as \texttt{ORG}. 
In case of DIPRE patterns, it is expected that {\em prefix}, {\em suffix} and {\em middle} strings of the patterns should match exactly. This condition hampers the coverage of the patterns. {\em SnowBall} patterns are designed to overcome this problem so that the minor variations in the text (e.g. misspellings, extra articles) do not cause a mismatch. Instead of having string patterns, {\em SnowBall} represents the context of entities (i.e. {\em prefix}, {\em suffix} and {\em right}) using word vectors in the vector space model. Higher the dot product between two context word vectors, higher is the similarity of the contexts. %with the help of the vector space model. 

\noindent\textbf{Evaluation of Patterns and Tuples:}
{\em SnowBall} discards all those patterns which are not precise enough, i.e. the patterns which are more likely to extract false positives. One way to discard such patterns is to filter out all the patterns not supported by some minimum number of seed examples. {\em SnowBall} also computes {\em confidence} for each pattern, but this computation assumes that one of the two named entities (say $NE_1$) is more important than the other. Suppose, the {\em confidence} of pattern $p$ is to be computed which extracts a candidate tuple $t_{curr}=(e_1, e_2)$ where $e_1$ is of type $NE_1$ and $e_2$ is of type $NE_2$. If there was a high confidence tuple $t_{prev} =(e_1, e'_2)$ generated in the previous iteration with the same entity ($e_1$) of type $NE_1$ as in $t_{curr}$, then this function compares entities $e_2$ and $e'_2$ which are of type $NE_2$. If these two are the same, then the tuple $t_{curr}$ is considered a {\em positive} match for the pattern $p$. Otherwise, it is considered as a {\em negative} match. The {\em confidence} of $p$ is then defined as follows:

{\footnotesize
\begin{equation}
Conf(p) = \frac{\#positive\_p}{\#positive\_p+\#negative\_p}
\end{equation}}
Here, $\#positive\_p$ and $\#negative\_p$ are the numbers of positive and negative matches for $p$, respectively. The confidence score of a candidate tuple is then computed using the confidence scores for patterns extracting it. %Consider a candidate tuple $t$ and set of patterns $P$ which extracted it. The {\em confidence} score of this tuple $t$ is computed as follows:
%{\small
%\begin{equation}
%Conf(t) = 1-\prod_{p\in P} \left( 1 - Conf(p)\cdot Match(p,c_p)\right)
%\end{equation}}
%Here, $c_p$ is the occurrence which was matched by the pattern $p$ while extracting the tuple $t$. 
Evaluation of patterns and tuples is the major differentiator between DIPRE and {\em SnowBall}, as low confidence patterns and tuples are discarded in each iteration avoiding most of the incorrect extractions. 

Most of the bootstrapping based techniques, apply relation patterns when both the entities are present as {\em name} mentions. % e.g. even if the system knows that \texttt{PER is son of PER} is a valid pattern to extract \texttt{PER-SOC} relation, this pattern won't be applied on the text such as \texttt{He is son of John.} Here, if the co-reference of \texttt{He} is resolved (say it is linked to a {\em name} \texttt{Bob}), then it is possible to extract \texttt{(Bob, John)} as a valid tuple of \texttt{PER-SOC} type. 
Gabbard et al.~\cite{gabbard2011coreference} explored the use of co-reference information for improving the performance (especially the recall) of the bootstrapped RE system. Other major bootstrapping approaches are by Ravichandran and Hovy~\cite{ravichandran2002learning}, Pantel and Pennacchiotti~\cite{pantel2006espresso}, Greenwood and Stevenson~\cite{greenwood2006improving}, Rosenfeld and Feldman~\cite{rosenfeld2007using}, Blohm and Cimiano~\cite{blohm2007using}, Xu et al.~\cite{xu2007seed}, Xu~\cite{xu2008bootstrapping}, Carlson et al.~\cite{carlson2009coupling} and Xu et al.~\cite{xu2010boosting}.

For mention-level RE i.e. ACE RDC task, Zhang~\cite{zhang2004weakly} proposed a bootstrapping based algorithm {\em BootProject} on top of SVMs. The SVM classifier uses the similar features used by the techniques discussed in the section~\ref{sec:featurebased}. They generalized the Co-training~\cite{cotraining1998} algorithm by relaxing following restrictions on multiple feature ``views'': mutual exclusivity, conditional independence and sufficiency for classification. 
%\begin{itemize}
%\item The feature views should be mutually exclusive and conditionally independent given the class label
%\item Each feature view should be sufficient for classification
%\end{itemize}
%Instead of explicitly splitting the feature space, Zhang~\cite{zhang2004weakly} generated multiple overlapping feature ``views'' by randomly selecting a subset of features from the original feature space. Multiple SVM classifiers are then trained initially using the seed examples corresponding to each ``view''. These classifiers then vote on the unlabelled instances and the instances with the highest agreement among the classifiers are then added to the initial seed examples' set. This process then repeats itself similar to other bootstrapping based approaches. 
Sun~\cite{sun2009two} observed that bootstrapping approaches like DIPRE and {\em SnowBall} are not good in extracting general relations like \texttt{EMP-ORG} relation in ACE 2004 dataset. 
%\texttt{EMP-ORG} relation actually consists of many subtypes like \texttt{executive}, \texttt{officer}, \texttt{soldier} etc. 
They proposed a two-stage bootstrapping approach where the first stage is similar to {\em SnowBall} whereas the second stage takes the patterns learned by the first stage as inputs and tries to extract more {\em nominals} (like \texttt{manager}, \texttt{CEO}, etc.) which indicate the general relation type \texttt{EMP-ORG}. Features based on such learned list of nominals are then incorporated into the supervised RE system for improved performance. The similar problem was addressed by Sun~\cite{sun2011semi} by using word clustering. In this approach, a large unlabelled corpus was used to learn word clusters, so that the words occurring in the similar context are grouped together in the same cluster. This is very useful to discover many new words (i.e. words not observed in the limited labelled data) which can be crucial to properly classify the relations. Features based on these word clusters were incorporated into the supervised RE system to achieve better performance. For mention level RE, Pawar et al.~\cite{pawar2013semi} proposed a semi-supervised approach using EM algorithm.

%Greenwood and Stevenson~\cite{greenwood2006improving} proposed a semi-supervised approach for learning of RE patterns. It is a bootstrapping based approach, but rather than starting with a set of seed entity pairs, it starts with a set of seed patterns which is grown over the iterations. The patterns are formed using the shortest path in the dependency tree connecting two entities.

It should be noted that the performance of bootstrapping based algorithm depends on the choice of initial {\em seed} examples. Analysis of quality of {\em seeds} chosen in bootstrapping algorithms is provided by Vyas et al.~\cite{vyas2009helping} and Kozareva and Hovy~\cite{kozareva2010not}.

\subsection{Active Learning}
Active Learning~\cite{settles2009active} techniques are being widely used by Machine Learning community in order to reduce annotation effort required to create labelled data. %Settles~\cite{settles2009active} provides a comprehensive survey of Active Learning techniques. 
The key idea behind active learning is that the learning algorithm is allowed to ask for true labels of some selected unlabelled instances. Various criterion have been proposed to choose these instances with the common objective of learning the underlying hypothesis quickly with a very few instances. The key advantage of active learning is that performance comparable with supervised methods is achieved through a very few labelled instances.

Sun and Grishman~\cite{sun2012active} presented an active learning system {\em LGCo-Testing}. It is based on an active learning approach Co-testing~\cite{cotesting2000} in the co-training~\cite{cotraining1998} setting. For applying Co-testing, 
%it is required to have two {\em uncorrelated} views of the instances such that each view is sufficient for labelling instances by itself. 
the authors proposed to create two views of the relation instances - i) a local view based on the features capturing the entity mentions being connected and other characteristics of the containing sentence; and ii) a global view based on the distributional similarity of the phrases connecting two entity mentions, using a large corpus. Suppose, for an instance of  \texttt{PHYS} type, the connecting phrase is \texttt{travelled to}, then examples of other phrases similar to it are \texttt{arrived in}, \texttt{visited}, etc. Distributional similarity would assign a high similarity between two phrases, if these phrases are observed in the similar context in a large corpus. A Maximum Entropy classifier is trained using the features from the local view. As a classifier using global view, a nearest neighbour classifier is used which uses the distributional similarity to find the nearest neighbours. 
%The {\em LGCo-testing} algorithm is described in detail in the table~\ref{tab:LGCotesting}. 
The {\em LGCo-testing} was further improved in terms of efficiency by Fu and Grishman~\cite{fu2013efficient}. Recently, a bilingual active learning approach for RE was proposed by Ya’nan et al.~\cite{acl2014bilingual} and the two languages were Chinese and English. Some of the other approaches using Active Learning for RE, are by Small and Roth~\cite{roth2008active} and Zhang et al.~\cite{zhang2012unified}.

%\begin{table}\footnotesize
%\begin{tabular}{p{\textwidth}}
%\hline
%\textbf{Input:} A few labelled instances $D_L$, Unlabelled instances $D_U$, Vector representations of all relation phrases occurring in $D_L$ as well as $D_U$\\
%\textbf{Output:} $D'_U$ (Gold-standard labels for some of the instances in $D_U$), Classifiers trained on $D_L \cup D'_U$\\
%1. Train a local classifier $C_{local}$ using $D_L$. A global classifier $C_{global}$ is a nearest neighbour classifier which uses $D_L$ as labelled instances and vector representations of relation phrases.\\
%2. Label the instances in $D_U$ using $C_{local}$ as well as $C_{global}$ to get a set $S$ of the relation instances for which $C_{local}$ and $C_{global}$ disagree.\\
%3. STOP if the set $S$ is empty.\\
%4. Pick a set of 5 most {\em informative} instances $S'$ from $S$ for obtaining labels from humans. Informativeness of an instance is calculated as the KL divergence between class probability distributions of two classifiers.\\
%5. $D_L:=D_L \cup S'$, $D_U:=D_U \ S'$\\
%6. Go back to step 1.\\
%\hline
%\end{tabular}
%\caption{LGCo-testing Algorithm}
%\label{tab:LGCotesting}
%\end{table}

\subsection{Label Propagation Method}
Label Propagation is a graph based semi-supervised method proposed by Zhu and Ghahramani~\cite{zhu2002learning} where labelled and unlabelled instances in the data are represented as nodes in a graph with edges reflecting the similarity between nodes. In this method, the label information for any node is propagated to nearby nodes through weighted edges iteratively and finally the labels of unlabelled examples are inferred when the propagation process is converged. The first attempt of using Label Propagation method for RE was by Chen et al.~\cite{chen2006relation}. They represented each entity pair (i.e. relation instance) in the dataset as a node in a graph and also associate a feature vector with it. The feature vector consists of various features characterizing the  relation instance as described in the section~\ref{sec:featurebased}. % for feature-based RE. 
The graph is completely connected with each edge between relation instances $R_i$ and $R_j$ having following weight,

{\small
\begin{equation*}
W_{ij} = \exp(\frac{s_{ij}}{\sigma^2})
\end{equation*}}
Here, $s_{ij}$ is the similarity between the feature vectors associated with $R_i$ and $R_j$. $\sigma^2$ is used to scale the weights, which the authors set to average similarity between labelled instances. Considering feature vector as probability distribution over features, the authors use JS divergence to compute distance between any two relation instances. Similarity between two instances is then inversely proportional to this distance. %JS divergence between two probability distributions $q$ and $r$ is defined as follows:
Chen et al.~\cite{chen2006relation} observed that this algorithm performed better than SVM and bootstrapping approaches. One of the major advantages of the label propagation is that the labels of the unlabelled instances are not only decided by the nearby labelled instances but also by the nearby unlabelled instances.

\subsection{Other Methods}
Jiang~\cite{jiang2009multi} applied multi-task transfer learning to solve a weakly-supervised RE problem. This special problem setting is that only a few seed instances of the relation type of interest are available but a large amount of labelled instances of other relation types is also available. The author observed that different relation types can share certain common structures, e.g. ACE relations EMP-ORG and GPE-AFF share the common syntactic structure where two entity mentions are connected through the preposition \texttt{of} (EMP-ORG : \texttt{employees \textbf{of} TCS}; GPE-AFF : \texttt{residents \textbf{of} India}). The proposed framework uses a multi-task transfer learning method along with human guidance in the form of entity type constraints. The commonality among different relation types is modelled through a shared weight vector, enabling the knowledge learned from other relation types to be transferred to the target relation type.

\subsection{Evaluation}
Some of the semi-supervised techniques focus on mention-level RE (e.g. ~\cite{chen2006relation,sun2009two,sun2011semi}) and these can be evaluated in the similar manner as that of supervised techniques given some labelled dataset is available. Most of the bootstrapping based techniques (like ~\cite{brin1999,agichtein2000snowball}) do not attempt to capture every mention of entity pairs, rather these techniques create a list of entity mention pairs exhibiting a particular relation type. While {\em precision} can be measured easily by verifying all the extracted pairs, it is difficult to estimate the {\em recall} as the number of true relation mentions in the unlabelled data is not available. In order to measure the {\em recall}, a smaller subset of unlabelled data can be considered and all the relation mentions within it can be manually identified. 

\section{Unsupervised Relation Extraction}
%Unsupervised RE techniques do not require any manually created, relation labelled data. %Even seed examples which are needed by semi-supervised techniques described in the previous section. 
In this section, we discuss some of the important unsupervised RE approaches which do not require any labelled data.
\subsection{Clustering based approaches}
One of earliest approaches for completely unsupervised RE was proposed by Hasegawa et al.~\cite{hasegawa2004discovering}. They only require a NER tagger to identify named entities in the text so that the system focuses only on those named entity mentions. The approach can be described in following steps:\\
%\begin{enumerate}
%\item 
1. The named entities in the text corpora are tagged\\
%\item 
2. Co-occurring named entity pairs are formed and their contexts are recorded\\
%\item 
3. Context similarities among the pairs identified in the step 2, are computed\\
%\item 
4. Using the similarity values computed in previous step, the pairs are clustered\\
%\item 
5. As each of these clusters represent one relation, a label is automatically assigned to each cluster describing the relation type represented by it
%\end{enumerate}
%We elaborate some of the important aspects of this approach.

\noindent\textbf{Named Entity (NE) pairs and context:} Two named entities are said to be co-occurring if there are at most $N$ intermediate words in between them. Pairs of all such co-occurring named entities are formed. All occurrences of a particular NE pair are observed and {\em all} the intermediate words for {\em all} such occurrences are recorded as the context for that NE pair. The words occurring to the left of first NE and the words occurring to the right of second NE are not considered to be part of the context. This is one of the limitations of this approach, as not all relations are expressed through using only intermediate words, e.g. \texttt{CEO of \textbf{ORG}, \textbf{PER} announced the financial results.} Also, the order of NEs is given importance, i.e. the pair $(NE_1, NE_2)$ is considered to be different than $(NE_2, NE_1)$ and their contexts are also recorded separately.

\noindent\textbf{Context similarity computation:} For each NE pair, a word vector is formed using all the words occurring in its context. Each word is weighted by $TF\times IDF$, where $TF$ is frequency of the word in the context and $IDF$ is inverse document frequency which is inversely proportional to number of NE pairs in whose context the word occurs. The authors use an interesting way to encode the order of NEs in the computation of $TF$. If a word $w$ occurs $L$ times in the context of $(NE_1, NE_2)$ and $M$ times in the context of $(NE_2, NE_1)$, then in the word vector for $(NE_1, NE_2)$, $TF$ of the word $w$ is computed as $L-M$. The intuition behind this is that this would be effective to detect the direction of a relation if the arguments of a relation have the same NE types. If the NE types of two NE pairs do not match, the similarity between the two pairs is considered to be $0$. The similarity of the contexts of two NE pairs is computed as the {\em Cosine Similarity} between their word vectors. The similarity value varies from $-1$ to $1$, where $1$ indicates that the contexts of two NE pairs are matching exactly and NEs occur in the same order. The similarity $-1$ indicates that the NE pairs have exactly the same context words but the order of NEs in them is reverse.

\noindent\textbf{Clustering and Labelling:} Using the similarity values, the NE pairs are clustered using hierarchical clustering with complete linkage. The resultant clusters are also labelled automatically using the high frequency words in the contexts of all the NE pairs in the cluster.

Chen et al.~\cite{chen2005unsupervised} proposed some improvements in Hasegawa et al.'s~\cite{hasegawa2004discovering} basic clustering approach. They developed an unsupervised feature selection method to remove uninformative noisy words from similarity computation. % for clustering.
%They also proposed a way to automatically select optimal number of clusters and employed a discriminative category matching (DCM) for auto-labelling the clusters by finding typical and discriminative words to represent different clusters.
Another similar approach for unsupervised RE from Wikipedia texts, was proposed by Yan et al.~\cite{yan2009unsupervised}. Here, instead of NE pairs, they form {\em Concept} pairs by using Wikipedia structure. For a Wikipedia article, its title becomes the {\em principal} concept and it is paired with other {\em secondary} concepts linking the current article to other Wikipedia articles. They proposed a two-step clustering approach from grouping the concept pairs with same relation type. In the first step, the concept pairs are clustered using similarity of the deep linguistic patterns linking the two concepts. These linguistic patterns are derived from the dependency trees of the containing sentences. Given the high quality of Wikipedia content, these patterns are usually more precise than the surface words context similarity computed in Hasegawa et al.~\cite{hasegawa2004discovering}. Once these highly precise clusters are formed in the first step, in order to improve the coverage, the second step of clustering is carried out on the remaining unclustered concept pairs. This clustering step uses the cluster centroids created in the first step and is based on simple surface patterns learned from larger Web corpus. %The authors observed that this combination of linguistic patterns and simple surface word patterns results in better performance.

Another interesting line of research %in the field of clustering based unsupervised RE
, is based on inducing relation types by generalizing dependency paths. Lin and Pantel~\cite{lin2001dirt} proposed DIRT (Discovery of Inference Rules) algorithm which is based on distributional similarity hypothesis. Rather than applying this hypothesis for discovering similar words, the authors use it to discover similar dependency paths which tend to link the same set of words. Poon and Domingos~\cite{poon2009unsupervised} presented a similar approach for USP (Unsupervised Semantic Parsing), which recursively clusters fragments of dependency trees such that various syntactic variations conveying the same meaning are grouped together. This idea of clustering expressions conveying the same meaning was extended to bilingual case by Lewis and Steedman~\cite{lewis2013unsupervised} for English and French. 
They learned the clusters of semantically similar English and French expressions representing some relations. Rather than using bilingual parallel corpus, the authors exploited the alignments between named entities in two languages. Another method for clustering of relation instances was proposed by Yao et al.~\cite{yao2011structured}. Their method makes use of generative probabilistic models, similar to LDA based topic models~\cite{blei2003lda}. These models represent relation instances in the form of entity mention types and various features based on connecting dependency path, which are generated by underlying hidden relation types. Here, the relation types play the role similar to underlying topics in the usual topic models for documents. The model also incorporates constraints between relation type and types of the entity mentions. The topic models proposed by Yao et al. were extended by de Lacalle and Lapata~\cite{de2013unsupervised} by integrating them with general domain knowledge. The domain knowledge is encoded as First Order Logic (FOL) rules which apply {\em Must-link} and {\em Cannot-link} constraints on the relation instances. %The intuition behind {\em Must-link} rules is that the relation instances sharing very important features should lie in the same cluster. Similarly, the intuition behind {\em Cannot-link} rules is that the relation instances which do not share any feature should never be clustered together.

\subsection{Other approaches}
One of the major non clustering based approach for unsupervised relation extraction is the URES (Unsupervised RE System) by Rosenfeld and Feldman~\cite{rosenfeld2006ures}. The only input required by the URES system is the definitions of the relation types of interest. A relation type is defined as a small set of keywords indicative of that relation type and entity types of its arguments. e.g. for the relation type {\em Acquisition}, the keywords can be \texttt{acquired, acquisition}. The URES system is a direct successor of the KnowItAll system~\cite{knowitall2005} which extracts facts from the web. The focus of KnowItAll is primarily on extracting entities and URES builds on that to extract relations. Feldman and Rosenfeld~\cite{feldman2006boosting} further boosted the performance of URES by introducing a simple rule based NER. Another interesting approach was proposed by Romano et al.~\cite{romano2006investigating} which is based on using unsupervised paraphrase acquisition for RE. %In simple terms, 
The text expressions that convey roughly the same meaning, are called as {\em paraphrases}. The approach begins with one text expression (and corresponding syntactic structure like dependencies structure) representing the target relation and finds its paraphrases using an unsupervised paraphrase acquisition approach. For example, starting from the initial expression \texttt{X interact with Y}, paraphrase acquisition algorithm would produce new expressions - \texttt{X bind to Y}, \texttt{X activate Y}, \texttt{X stimulate Y}, \texttt{interaction between X and Y}, etc. %These paraphrases i.e. new text expressions are expected to convey the same meaning as that of the initial expression and hence can be used to extract more instances of the target relation. 

\section{Open Information Extraction}
Traditional RE focuses on precise, pre-specified set of relations. Extensive human involvement is generally required in designing extraction rules or for creating labelled training data. Hence, it is difficult make such systems work in a different domain. To overcome these limitations, the paradigm of Open Information Extraction (Open IE) was first proposed by Banko et al.~\cite{banko2007}, in the form of the TextRunner system. Open IE systems automatically discover possible relations of interest using the text corpus without any human involvement. Hence, no additional efforts are required to switch to a different domain. %It is scalable and efficient and hence can work with web-scale heterogeneous corpus. %In this section, we describe two such Open IE systems TextRunner~\cite{banko2007} and ReVerb~\cite{fader2011}.

%\subsection{TextRunner}
The TextRunner system~\cite{banko2007} consists of following three core modules:\\
%\begin{enumerate}
%\item 
\textbf{1. Self-supervised Learner:} Using some heuristic rules, it automatically labels a set of extracted entity tuples as positive or negative. Here, positive class indicates that the corresponding tuple represents some valid relation.
%It is so called because it automatically labels its own training data as positive (trustworthy) or negative. A small sample corpus is selected and %syntactic structures 
%parse trees for all sentences are obtained. For each parsed sentence, all base noun phrases are marked as entities.
%For each pair of entities in a sentence, extractions are made in the form of tuple $(E_1, R, E_2)$ : $E_1$ and $E_2$ are strings that denote two entities, $R$ is formed by traversing parse structure connecting $E_1$ to $E_2$. The Learner labels each tuple as positive or negative depending on certain conditions. These conditions try to capture whether any meaningful relation exists between the entities. Some of the heuristics used for labelling any tuple as positive are as follows:\\
%\begin{table}[h]
%\begin{tabular}[h]{l}
%\hspace{0.4cm}1. Dependency chain between $E_1$ and $E_2$ is not too long\\
%\hspace{0.4cm}2. The path from $E_1$ to $E_2$ tree does not cross a sentence-like boundary\\
%\hspace{0.4cm}3. Neither $E_1$ nor $E_2$ consist solely of a pronoun\\
%\end{tabular}
%\end{table}
%\begin{itemize}
%\item 
%\hspace{0.4cm}1. Dependency chain between $E_1$ and $E_2$ is not too long\\
%\item 
%\hspace{0.4cm}2. The path from $E_1$ to $E_2$ tree does not cross a sentence-like boundary\\
%\item 
%\hspace{0.4cm}3. Neither $E_1$ nor $E_2$ consist solely of a pronoun\\
%\end{itemize}
After automatic labelling, each tuple is mapped to a feature vector representation %. %Some of the features used are - POS tag sequence in relation string $R$, Number of tokens, stopwords in $R$, Whether $E_1$ or $E_2$ is a proper noun, POS tag left of $E_1$, POS tags right of $E_2$.
%\begin{itemize}
%\item POS tag sequence in relation string $R$
%\item Number of tokens, stopwords in $R$
%\item Whether $E_1$ or $E_2$ is a proper noun
%\item POS tag left of $E_1$, POS tags right of $E_2$
%\end{itemize}
%It is important to note here that none of the features are based on parsing information. Hence, sentences need not be parsed for deciding which of the tuples contained within them are trustworthy. A 
 and a Naïve Bayes classifier is trained.\\% using the tuples with above feature representations.\\
%\item 
\textbf{2. Single Pass Extractor:} It makes a single pass over entire corpus and obtains POS and NP (base noun phrases) information for all sentences. For each sentence, each pair of NPs ($E_1$ and $E_2$) becomes a candidate tuple and the corresponding relation string $R$ is found by examining the text in between. For each word occurring in between, it is heuristically decided whether to include it in $R$. 
%A sequence-labelling graphical model like CRF can be used for this purpose. 
%Some non-essential words are removed heuristically, e.g. \texttt{definitely developed} is reduced to \texttt{developed}. 
Each candidate tuple is presented to the Naïve Bayes classifier and only those tuples which are classified as ``positive'' are extracted and stored.\\
%\item 
\textbf{3. Redundancy-based Assessor:} After extractions are performed over entire corpus, TextRunner automatically merges some of the tuples where both the entities and relations are identical. For each tuple, number of distinct sentences containing it is also recorded and the assessor then uses these counts to assign a probability of correctness to each tuple. %Assessor estimates the probability that a tuple $(E_1, R, E_2)$ is a correct instance of the relation $R$ between $E_1$ and $E_2$ given that it was extracted from $k$ different sentences.

Banko et al.~\cite{banko2008tradeoffs} proposed to use Conditional Random Field based, self-supervised sequence classifier O-CRF instead of Naive Bayes classifier used in TextRunner and observed better performance. 
%They observed that O-CRF outperformed Naive Bayes considerably. 
Another improvement to TextRunner was suggested by Wu and Weld~\cite{wu2010open} in their Wikipedia-based Open Extractor (WOE) system. They used Wikipedia infoboxes to more accurately generate training data for the Self-supervised Learner module. Similar approach of using Wikipedia infoboxes was adopted by Weld et.al.~\cite{weld2009using} in their {\em Kylin} open IE system. Bootstrapping methods like Snowball~\cite{agichtein2000snowball} significantly reduce the number of initial training examples, but these methods do not perform open IE. Zhu et al.~\cite{zhu2009statsnowball} proposed a bootstrapping approach {\em StatSnowball} which can even perform open IE along with traditional RE.

%\subsection{ReVerb}
Fader et al.~\cite{fader2011} proposed {\em ReVerb}, an advanced Open IE system which improves over TextRunner by overcoming following limitations of TextRunner:\\% of uninformative and incoherent extractions often faced in Open IE by proposing some syntactic and lexical constraints to improve extraction quality.
%Major limitations of some of the early Open IE systems including TextRunner were as follows:\\
%\begin{enumerate}
%\item 
\textbf{1. Incoherent Extractions:} No meaningful interpretation of extracted relation phrases can be made. %, are termed as {\em Incoherent Extractions}.% (see Table~\ref{tab:incoherent_extractions_examples} for examples). 
 Such extractions are result of a word-by-word decision making about whether to include a word in a relation phrase.\\
%\begin{table}\footnotesize
%\centering
%\begin{tabular}{|p{7.5cm}|c|}
%\hline
%\textbf{Sentence} & \textbf{Incoherent Relation} \\
%\hline
%\texttt{The guide \underline{contains} dead links and \underline{omits} sites.} & \texttt{contains omits}\\
%\hline
%\texttt{The Mark 14 \underline{was central} to the \underline{torpedo} scandal of the fleet.} & \texttt{was central torpedo}\\
%\hline
%\texttt{They \underline{recalled} that Nungesser \underline{began} his career as a precinct leader.} & %\texttt{recalled began}\\
%\hline
%\end{tabular}
%\caption{Examples of Incoherent Extractions}
%\label{tab:incoherent_extractions_examples}
%\end{table}
\textbf{2. Uninformative Extractions:} These extractions omit critical information and are generally caused by improper handling of relation phrases that are expressed by Light Verb Constructions (LVCs). %are one of the important types of constructions that are missed by TextRunner. 
LVCs are multi-word expressions composed of a verb and a noun, with the noun carrying semantic content. e.g. \texttt{is the author of}. %, \texttt{has a cameo in}, \texttt{gave birth to}. 
From the sentence \texttt{John made a promise to Alice}, the TextRunner makes an uninformative extraction \texttt{(John, made, a promise)} whereas correct extraction is \texttt{(John, made a promise to, Alice)}.\\
%\item 
\textbf{3. Overly-specific Extractions:} TextRunner may extract very specific relation phrases which are not useful. %For example, consider the sentence: \texttt{The Obama administration is offering only modest greenhouse gas reduction targets at the conference.} 
e.g. \texttt{(The Obama administration, is offering only modest greenhouse gas reduction targets at, the conference)}% which has a overly-specific relation phrase.
%\end{enumerate}

To overcome the above limitations, the {\em ReVerb} algorithm proposes following two constraints on relation phrases to be extracted. %The first one is a {\em syntactic constraint} which eliminates incoherent extractions and also reduces uninformative extractions. The second constraint is a {\em lexical constraint} which eliminates most of the overly-specific extractions.

%\subsubsection{Syntactic Constraint}
\noindent\textbf{Syntactic Constraint:} The relation phrases are constrained to match the POS tag pattern mentioned in the Table~\ref{tab:syntactic_constraints}. This constraint limits relation phrases to be either one of following:
%\begin{table}[h]
%\begin{tabular}{p{\textwidth}}
%\hspace{0.1cm}1. A verb (e.g. \texttt{invented})\\
%\hspace{0.1cm}2. A verb immediately followed by a preposition (e.g. \texttt{born at})\\
%\hspace{0.1cm}3. A verb followed by nouns, adjectives or adverbs ending in preposition (e.g. \texttt{has atomic weight of})\\
%\hspace{0.1cm}4. Multiple adjacent matches are merged into a single relation phrase (e.g. \texttt{wants to extend})\\
%\end{tabular}
%\end{table}
%\begin{itemize}
%\item 
a verb (e.g. \texttt{invented}); a verb immediately followed by a preposition (e.g. \texttt{born at}); a verb followed by nouns, adjectives or adverbs ending in preposition (e.g. \texttt{has atomic weight of}); multiple adjacent matches merged into a single relation phrase (e.g. \texttt{wants to extend})
%\end{itemize}
\begin{table}[h]\footnotesize\center
%\centering
\begin{tabular}{|c|}
\hline
%{\large $V|VP|VW^{*}P$}\\
$V|VP|VW^{*}P$\\
%{\large $V=verb$ $particle?$ $adv?$}\\
$V=verb$ $particle?$ $adv?$\\
%{\large $W=(noun|adj|adv|pron|det)$}\\
$W=(noun|adj|adv|pron|det)$\\
%{\large $P=(prep|particle|inf.$ $marker)$ }\\
$P=(prep|particle|inf.$ $marker)$\\
\hline
\end{tabular}
\caption{Syntactic Constraint}
\label{tab:syntactic_constraints}
\end{table}
\vspace{-3mm}
Incoherent extractions are avoided because there are no isolated word-level decisions about whether to include a word in relation phrase. The decision is taken for a sequence of words whether that sequence satisfies the POS pattern. Uninformative extractions are avoided because nouns are also allowed as a part of relation phrase and relations expressed via LVCs are also captured.

%\subsubsection{Lexical Constraint}
\noindent\textbf{Lexical Constraint:} %There are some phrases which satisfy the syntactic constraint but do not express any relation. e.g. \texttt{monitor the human rights situation in}. 
To avoid overly-specific relation phrases, a lexical constraint is applied that considers only those relation phrases as valid which take at least $k$ distinct argument pairs. %The intuition behind it is : a valid relation phrase should take many distinct arguments in a large corpus. 
A valid relation phrase like \texttt{took control over} occurs with multiple distinct arguments like \texttt{(Germany, took control over, Austria)} and \texttt{(Modi, took control over, administration)}. 
%Hence, only those relation phrases are considered valid which take at least $k$ distinct argument pairs. %At the web scale, $k=20$ was observed to work well.

%The loss in recall due to the above syntactic and lexical constraints was observed to be around $15\%$. Some of the important types of valid relation phrase types which are missed due to the constraints are:\\
%\begin{itemize}
%\item 
%-\textbf{Non-contiguous phrase structures} like \texttt{\textbf{is produced} and maintained \textbf{by}} and \texttt{\textbf{founded} in 1995 \textbf{by}}\\
%\item 
%-\textbf{Relation phrase not between arguments} like \texttt{\textbf{directed by} Spielberg, Jurassic Park....}\\
%\item 
%-\textbf{POS pattern does not match} for phrases like \texttt{has a lot of faith in}
%\end{itemize}

%The {\em ReVerb} algorithm extracts tuples in two phases:
%\begin{itemize}
%\item \textbf{Relation Extraction:} Relation phrases are identified satisfying syntactic and lexical constraints explained earlier.
%\item \textbf{Argument Extraction:} A pair of arguments $(E_1,E_2)$ are found for each relation phrase $r$, resulting in extraction of tuple $(E_1, r, E_2)$. $E_1$ is the first noun phrase occurring to the left of $r$ and $E_2$ is the first noun phrase occurring to the right of $r$.
%\end{itemize}
%To summarize, 
{\em ReVerb} differs from TextRunner in the manner in which the relation phrases are identified. Relation phrases are identified ``holistically'' rather that the word-by-word decision in TextRunner, resulting in more meaningful relation phrases. %Also the relation phrases are filtered as per the lexical constraint. 
{\em ReVerb} follows a ``relations first'' approach rather than TextRunner's ``arguments first'' approach, as it first identifies a valid relation phrase and then extracts the arguments. This helps not to confuse a noun in relation phrase as an argument. e.g. {\tt promise} in {\tt made a promise to}. %The resulting extractions are also assigned confidence scores using a logistic regression classifier.

Etzioni et al.~\cite{etzioni2011open} observed that almost 65\% of {\em ReVerb}'s extraction had a correct relation phrase but an incorrect arguments. They proposed an Open IE system R2A2 which is an improvement over {\em ReVerb} and contains an additional module {\em ArgLearner} for identifying arguments. %After linguistic analysis of these errors, they found that simple NPs (like \texttt{Obama}, \texttt{vegetable oils}, etc.) do not cover all the cases for arguments. In order to improve the coverage, other cases need to be considered which are as follows. The arguments in focus are underlined and the relation phrases are italicized.
%\begin{itemize}
%\item NP with attached prepositional phrases, e.g. \texttt{\underline{The forest in Brazil} \textit{is threatened by} ranching.}
%\item Lists, e.g. \texttt{\underline{Infosys and Wipro} \textit{are headquartered in} Bangalore.}
%\item Independent clause, e.g. \texttt{World Bank \textit{predicts} \underline{that India's GDP growth} \underline{rate will be more that 7.5\%}}
%\item Relative clause, e.g. \texttt{\underline{Mumbai, which is located in India}, \textit{has a population of} 15 million.}
%\end{itemize}
%The {\em ArgLearner} module of R2A2 employs three classifiers to identify the arguments, given a relation phrase and the corresponding sentence. Two classifiers are used to detect the start and end of the left argument and one classifier is used to detect the end of the right argument. It was observed that the right argument occurs immediately to the right of the relation phrase in almost all the cases, hence there was no need to build a separate classifier for that.
{\em ReVerb} is found to outperform all the previous open IE systems like TextRunner and WOE. And R2A2 was observed to achieve better precision and recall than even {\em ReVerb}. In order to be efficient on the web scale, most of the open IE systems do not perform deep syntactic analysis. Gamallo~\cite{gamallo2012dependency} used robust and fast dependency parsing in their open IE system {\em DepOE} on the Web scale to achieve more precise extraction than {\em ReVerb}. Mesquita et al.~\cite{mesquita2013effectiveness} presented a comprehensive comparison of $8$ Open Information Extraction techniques for their efficiency and effectiveness. They analysed the trade-off between the complexity of NLP processing (i.e. from simpler POS tagging to more complex Semantic Role Labelling) versus effectiveness. Some of the major advantages of Open IE systems are their unsupervised nature and scalability to the Web scale. Recently, Open IE has been an active area of research within RE systems. One of the major limitation of these systems is that the same semantic relation may be represented by multiple relation phrases and some post-processing is required to consolidate such various representations of the same relation type.

\section{Distant Supervision}
Distant Supervision, proposed by Mintz et al.~\cite{mintz2009}, is an alternative paradigm which does not require labelled data. The idea is to use a large semantic database for automatically obtaining relation type labels. Such labels may be noisy, but the huge amount of training data is expected to offset this noise. Similar ideas of creating ``weakly'' labelled training data, were earlier proposed by Craven and Kumlien~\cite{craven1999constructing}, Bunescu and Mooney~\cite{bunescu2007learning} and Nguyen et al.~\cite{nguyen2007exploiting}. Distant Supervision combines advantages of both the paradigms : Supervised and Unsupervised. It combines thousands of features using a probabilistic classifier as in the case of supervised paradigm. Also, it extracts a large number of relations from large corpora of any domain as in the case of unsupervised paradigm. Mintz et al.~\cite{mintz2009} used Freebase~\cite{kurt2008freebase} as a semantic database which stores pairs of entities for various relations.\\%, see the Table~\ref{tab:freebase_examples} for some examples.
%\begin{table}\footnotesize
%\centering
%\begin{tabular}{|c|c|}
%\hline
%Relation Name & Example \\
%\hline
%/people/person/nationality & \texttt{John Dugard, South Africa} \\
%\hline
%/location/location/contains & \texttt{Belgium, Nijlen} \\
%\hline
%/people/person/profession & \texttt{Dusa McDuff, Mathematician} \\
%\hline
%/people/person/place\_of\_birth & \texttt{Edwin Hubble, Marshfield} \\
%\hline
%\end{tabular}
%\caption{Some example entries from Freebase}
%\label{tab:freebase_examples}
%\end{table}
\noindent\textbf{Labelling heuristic:} If two entities participate in a relation, any sentence that contains both of them might express that relation. For example, Freebase contains entity pair \texttt{<M. Night Shyamalan, The Sixth Sense>} for the relation /film/director/film, hence both of the following sentences are considered to be positive examples for that relation:\\
%\begin{enumerate}\small
%\item 
{\footnotesize
1. \texttt{\underline{M. Night Shyamalan} gained international recognition when he wrote and directed 1999's \underline{The Sixth Sense}.}\\
%\item 
2. \texttt{\underline{The Sixth Sense} is a 1999 American supernatural thriller drama film written and directed by \underline{M. Night Shyamalan}.}}
%\end{enumerate}

\noindent\textbf{Negative Instances:} The above mentioned heuristic can be used to obtain only positive instances for various relation types but not the negative instances. In order to train a classifier both the types of instances are necessary. Entity pairs which do not appear in any Freebase relation are randomly selected and treated as negative instances. Some entity pairs may be incorrectly labelled as negative due to incompleteness of Freebase.

Using the automatically obtained labelled data, a multi-class Logistic Classifier with Gaussian regularization is trained. % in an usual supervised manner. 
Various lexical, syntactic and named entity type features are used for training. %Lexical features include the sequence of words between the two entities, the part-of-speech tags of these words, a flag indicating which entity came first in the sentence, a window of $k$ words to the left (right) of Entity 1 (Entity 2) and their part-of-speech tags. Syntactic features are conjunctions of a dependency path connecting two entities and one window node that is not part of dependency path. Window node is a node connected to one of the two entities and not part of the dependency path. Feature conjunctions are used instead of using the features independently. This results in low recall but high precision features. As the training data size is quite large even high precision complex features occur multiple times.
There were several subsequent efforts to improve upon the approach by Mintz et al.~\cite{mintz2009} like Yao et al.~\cite{yao2010collective}, Hoffmann et al.~\cite{hoffmann2010learning}, Reidel et al.~\cite{riedel2010modeling}, Nguyen and Moschitti~\cite{nguyen2011end}, Takamatsu et al.~\cite{takamatsu2012reducing} and  Krause et al.~\cite{krause2012large}.

One of the major shortcoming of the traditional distant supervision based approaches was that they failed to model overlapping relations, i.e. the fact that for the same pair of entities, there can be multiple valid relations. e.g. \textit{FoundedBy}\texttt{(Steve Jobs, Apple)} and \textit{CEO}\texttt{(Steve Jobs, Apple)}. Two of the major approaches to handle this problem were proposed by Hoffmann et al.~\cite{hoffmann2011knowledge} and Surdeanu et al.~\cite{surdeanu2012multi}. The Multi-instance Multi-label learning based approach (MIML-RE) by Surdeanu et al.~\cite{surdeanu2012multi}, models latent relation labels for multiple instances (occurrences) of an entity pair. It also models dependencies between labels for a single entity pair.% are assigned to multiple instances of entity pairs, hence the term {\em Multiple instances}. It also captures dependencies between labels for a single entity pair, hence the term {\em Multiple labels}.

MIML-RE uses a novel graphical model for representing ``multiple instances'' as well as ``multiple labels'' of an entity pair. A mention level relation classifier is used to identify relation label for each {\em mention} of an entity pair using features derived from the context of the mention. There is another set of classifiers, one per each distinct relation label, which operate at the entity pair level. These are binary classifiers indicating whether a specific relation holds for an entity pair. %These classifiers help to capture information that cannot be modelled by the mention level classifier. e.g. 
These classifiers can learn that two relation labels like {\em BornIn} and {\em SpouseOf} cannot be generated jointly for the same entity pair. If mention level classifier in the lower layer assigns both of these labels for different mentions of the same tuple, then one of them can be cancelled by the Entity pair level classifiers. It can also learn when the two labels tend to appear jointly, like {\em CapitalOf} and {\em ContainedIn}. In order to learn various parameters of this graphical model, hard discriminative EM algorithm is used.

MIML-RE outperforms not only the traditional distant supervision approach by Mintz et al.~\cite{mintz2009} but also the approach modelling multiple instances by Hoffmann et al.~\cite{hoffmann2011knowledge}. One of the major advantage that MIML-RE has over the Hoffmann et al.~\cite{hoffmann2011knowledge} %which uses a deterministic ``at least one'' decision, 
is its entity pair level classifiers. The datasets used for evaluation are Riedel's dataset~\cite{riedel2010modeling} and KBP dataset. The KBP dataset was constructed by Surdeanu et al.~\cite{surdeanu2012multi} using the resources distributed for the 2010 and 2011 KBP shared tasks~\cite{kbp2010,kbp2011}. These datasets are widely used in the distant supervision related literature. 

Recently, distant supervision for RE has become a very active field of research with several new approaches to overcome some of the specific problems. Due to incompleteness of semantic database used for labelling by distant supervision, many {\em negative} instances are actually false negatives. To overcome this problem, Min et al.~\cite{min2013distant} proposed an algorithm which learns only from positive and unlabelled examples. Xu et al.~\cite{xu2013filling} addressed this problem of false negative training examples by adapting the information retrieval technique of pseudorelevance feedback. Zhang et al.~\cite{zhang2012big} analyzed comparative performance of distant supervision and crowd-sourcing which is also an alternative low-cost method of obtaining labelled data. Pershina et al.~\cite{pershina2014infusion} proposed a {\em Guided DS} approach which shows that when small amount of human labelled data is available along with distantly labelled, a significant improvement in RE performance is observed. %They extended the MIML-RE model to systematically incorporate human labels as training guidelines represented as conjunctions of some general but high confidence features. 
One problem with MIML-RE's data likelihood expression is that it is a non-convex formulation. Grave~\cite{grave2014convex} proposed a new approach based on discriminative clustering which leads to a convex formulation. 
%Koch et al.~\cite{koch2014type} investigated various improvements such as integrating named entity linking (NEL) and co-reference resolution, enforcing type constraints of linked arguments and partitioning the model by relation type signature. Liu~\cite{liu2014exploring} also proposed a novel method for exploring entity type constraints and integrating them with the RE model. 
Other recent approaches with various improvements over the basic distant supervision approach, are proposed by Zhang et al.~\cite{zhang2013towards}, Chen et al.~\cite{chen2014encoding}, Han and Sun~\cite{han2014semantic}, Nagesh et al.~\cite{nagesh2014noisy}, Liu~\cite{liu2014exploring} and Koch et al.~\cite{koch2014type}.

Distant supervision can not be applied if the relation of interest is not covered explicitly by the knowledge base (like FreeBase). Zhang et al.~\cite{zhang2012ontological} proposed a novel approach of {\em Ontological Smoothing} to address this problem when at least some seed examples of the relation of interest are available. {\em Ontological Smoothing} generates a mapping between the relation of interest and the knowledge-base. %Let the relation of interest be {\em IsCoachedBy(PLAYER,COACH)} and the knowledge base does not contain this relation explicitly, rather it contains other relations such as {\em PlaysFor(PLAYER,TEAM)} and {\em TeamCoach(TEAM,COACH)}. {\em Ontological Smoothing} would find the mapping of {\em IsCoachedBy} to the knowledge base through the ``join'' of two relations {\em PlaysFor} and {\em TeamCoach}. 
Such mappings are used to generate additional training examples along with seed examples and distant supervision is then used to learn the relation extractor.
An interesting study carried out by Nguyen et al.~\cite{nguyen2011joint}, reported that joint distant and direct supervision can significantly improve the RE performance as compared to the systems which use only gold-standard labelled data like ACE 2004. They mapped some YAGO~\cite{suchanek2007yago} relation types to the seven ACE 2004 relation types and created a distantly supervised labelled dataset using Wikipedia text. Two separate relation classifiers were trained - one using only ACE 2004 labelled data and other using distantly supervised labelled data combined with ACE 2004 data. The linear combination of the probabilities obtained from both of these classifiers was considered as the final probability. This joint classifier provided around $3$\% improvement in F-measure compared to the classifier using only ACE 2004 training data.

A first attempt of using Active Learning with distantly supervised RE, was reported by Angeli et al.~\cite{angeli2014combining}. They proposed to provide a partial supervision to MIML-RE with the help of active learning. A novel selection criteria was proposed for selecting relation instances for labelling. This criteria prefers relation instances which are both uncertain (high disagreement in committee of classifiers) and representative (similar to large number of unlabelled instances). The annotations were obtained through crowdsourcing from Amazon Mechanical Turk and a significant improvement over MIML-RE was observed.

\section{Recent Advances in Relation Extraction}
In this section, we describe some recent advances in the field of RE.% which do not exactly fit in any of the techniques explained before.

\noindent\textbf{Universal Schemas:} Riedel et al.~\cite{riedel2013relation} proposed to use Universal Schemas, which are union of relation types of existing structured databases (e.g. FreeBase, Yago) and all possible relation types in the form of surface forms used in Open IE. They proposed an approach to learn asymmetric implicature among these universal relation types. This implicature helps to infer possible new relations between an entity pair given a set of existing relations for that entity pair from some structured database. If a city and a country are known to be related with a relation type {\em CapitalOf}, it can be inferred that the relation type {\em LocatedIn} also holds true for that pair, but not vice versa as the implicature is asymmetric.
%A matrix is constructed with number of columns equal to the total number of relation types in the Universal Schema and number of rows equal to the total number of entity pairs. Any cell $(r,p)$ in this matrix is supposed to represent the probability that the relation type $r$ holds true for the entity pair $p$. The matrix can be initialized with the observed facts, i.e. all the cells $(r,p)$ where the relation type $r$ is known to exist between entities of pair $p$ (as per the FreeBase or any other database), are initialized to $1$. This situation is analogous to the matrix of customers and products where each cell $(p,c)$ represents the probability the customer $c$ likes the product $p$. Hence, the authors borrow the collaborative filtering techniques~\cite{su2009survey} like Matrix Factorization and Implicit Feedback to solve the problem of relations implicature. 
Other similar approaches were proposed by Chang et al.~\cite{chang2014typed} and Fan et al.~\cite{fan2014distant}.
%In another similar work, Chang et al.~\cite{chang2014typed} proposed a tensor decomposition approach for knowledge base embedding which is useful for discovering new relations missing from existing databases.

\noindent\textbf{n-ary Relation Extraction:} The relations among more than two entities are generally referred to as {\em Complex} or {\em Higher Order} or {\em n-ary} relations. Example of an n-ary relation is \texttt{EMP-ORG-DESG} which represents relation between a person, the organization where he/she is employed and his/her designation. This relation exists for entities \texttt{(John Smith, CEO, ABC Corp.)} in the sentence : \texttt{John Smith is the CEO of ABC Corp.} %Another such n-ary relation is \texttt{GIFT(Singh, Obama, painting)} from the sentence : \texttt{Indian Prime Minister Singh gifted a paining to American President Obama.} 
One of the earliest attempt to address this problem was by McDonald et al.~\cite{mcdonald2005simple}. They used well-studied binary RE to initially find relations between all possible entity pairs. The output of this binary RE was represented as a graph where entities correspond to nodes and valid relations between entities correspond to edges. The authors then proposed to find maximal cliques in this graph such that each clique corresponds to some valid n-ary relation. They demonstrated effectiveness of this approach on the Biomedical domain dataset. Another recent approach for n-ary RE is an Open IE system named Kraken proposed by Akbik and L{\"o}ser~\cite{akbik2012kraken}. Zhou et al.~\cite{zhou2014biomedical} surveyed several complex RE approaches in the biomedical domain.

Another perspective to look at n-ary RE problem as Semantic Roles Labelling (SRL). The SRL task is to identify predicate and its arguments in a given sentence automatically. One of the fundamental SRL approach is by Gildea and Jurafsky~\cite{gildea2002automatic} and some standard semantic roles labelled datasets are PropBank~\cite{kingsbury2002treebank} and FrameNet~\cite{baker1998berkeley}.

\noindent\textbf{Cross-sentence Relation Extraction:} Most of the techniques that we have discussed, focus on intra-sentential RE, i.e. entity mentions constituting a relation instance occur in the same sentence. Swampillai and Stevenson~\cite{swampillai2011extracting} proposed an approach to extract both intra-sentential and inter-sentential relations. Some examples of an inter-sentential relation is shown in the Table~\ref{tab:examples_cross_sentence}.
\begin{table}[h]\footnotesize
\begin{tabular}{|c|p{9cm}|c|}
\hline
\textbf{No.}&\textbf{Sentences} & \textbf{Relation}\\
\hline
\multirow{2}{*}{1} & \texttt{The youngest son of ex-dictator \underline{Suharto} disobeyed a summons to surrender himself to prosecutors Monday and be imprisoned for corruption.} & \multirow{2}{*}{PER-SOC}\\
\cline{2-2}
 & \texttt{\underline{Hutomo ``Tommy'' Mandala Putra}, 37, was sentenced to 18 months in prison on Sept. 22 by the Supreme Court, which overturned an earlier acquittal by a lower court.} & \\
\hline
\multirow{2}{*}{2} & \texttt{Computer intruders broke into \underline{Microsoft Corp.} and were able to view some of the company's source code, the basic program instructions, for a future software product, the company said Friday. } & \multirow{2}{*}{EMP-ORG}\\
%\cline{2-2}
% & {\footnotesize \texttt{But the unknown culprits, who had access to some of the company's computers for an undetermined period, were not able to view or steal the company's crucial source code for its Windows or Office software, a company spokesman said Friday afternoon. }} & \\
\cline{2-2}
 & \texttt{``The situation appears to be narrower than originally thought,'' said the spokesman, \underline{Mark Murray}.} & \\ 
\hline
\end{tabular}
\caption{Examples of relations spanning across two consecutive sentences. The entity mentions between which the relation holds, are underlined.}%shown in \texttt{\textbf{BOLD}} font}
\label{tab:examples_cross_sentence}
\end{table}
\vspace{-3mm}
The authors adapted the structured features (like parse tree paths) and techniques for intra-sentential RE for the inter-sentential situation. Generally, it can be observed that most of the cases (like Example 1 in table~\ref{tab:examples_cross_sentence} but not Example 2) of inter-sentential RE can be addressed through co-reference resolution~\cite{elango2005coreference}. In Example 1, \texttt{son} in the first sentence actually refers to \texttt{Hutomo ``Tommy'' Mandala Putra} in the second sentence. Given that intra-sentential RE technique can detect \texttt{PER-SOC} relation between \texttt{son} and \texttt{Suharto}, using the co-reference we can get the required inter-sentence relation.

\noindent\textbf{Convolutional Deep Neural Network:} Zeng et al.~\cite{zeng2014relation} explored the feasibility of performing RE without any complicated NLP preprocessing like parsing. They employed a convolutional DNN to extract lexical and sentence level features. They observed that the automatically learned features yielded excellent results and can potentially replace the manually designed features that are based on the various pre-processing tools like syntactic parser. Other similar techniques which use Recursive Neural Networks were proposed by Socher et al.~\cite{socher2012semantic} and Hashimoto et al.~\cite{hashimoto2013simple}.

\noindent\textbf{Cross-lingual Annotation Projection:} Entities and relations labelled data is available only for a few {\em resource-rich} languages like English, Chinese and Arabic. Kim et al.~\cite{kim2010cross} proposed a technique to project relation annotations from a {\em resource-rich} source language (English) to a {\em resource-poor} target language (Korean) by utilizing parallel corpus. Direct projection was used in the sense that the projected annotations were determined in a single pass by considering only alignments between entity candidates. Kim and Lee~\cite{kim2012graph} proposed a graph-based projection approach which utilizes a graph that is constructed with both entities and context information and is operated in an iterative manner.

\noindent\textbf{Domain Adaptation:} The fundamental assumption of the supervised systems is that the training data and the test data are from the same distribution. But when there is a mismatch between these data distributions, the RE performance of supervised systems tends to degrade. This generally happens when supervised systems are used to classify out-of-domain data. In order to address this problem, domain adaptation techniques are needed. The first such study for RE was carried out by Plank and Moschitti~\cite{plank2013embedding}. They reported that the out-of-domain performance of kernel-based systems can be improved by embedding semantic similarity information obtained from word clustering and latent semantic analysis (LSA) into syntactic tree kernels. Nguyen and Grishman~\cite{nguyen2014employing} proposed an adaptation approach which generalizes lexical features using both word cluster and word embedding~\cite{bengio2001neural} information. Another approach by Nguyen et al.~\cite{nguyen2014robust} proposed to use only the relevant information from multiple source domains which results in accurate and robust predictions on the unlabelled target-domain data.

\section{Conclusion and Future Research Directions}
To the best of our knowledge, we presented a first comprehensive survey of relation extraction techniques. We clarified the usage of the term ``Relation Extraction'' which can refer to either mention-level RE (ACE RDC task) and global RE. We first described supervised techniques including important feature-based and kernel based techniques. We discussed how these techniques evolved over a period of time and how they are evaluated. We observed that among all the supervised techniques, syntactic tree kernel based techniques were the most effective. They produced the best results either when combined with some other kernel to form a composite kernel or when dynamically determined tree span was used. The best reported result after almost a decade of efforts on the ACE 2004 dataset is around $77$\%. Hence, we think that there is still some room for improvement here. 

We also covered joint modelling techniques which jointly extract entity mentions and relations between them. From a practical point of view, this problem is quite important because good entity extraction performance is a must for achieving good RE performance. The joint modelling techniques allows two-way information flow between these tasks and try to a achieve better performance for both as compared to isolated models. We then focussed on some of the important semi-supervised and unsupervised techniques. There has been a lot of work in these areas, as increasing amount of efforts are being put to reduce the dependence on the labelled data. We also covered the paradigms of Open IE and Distant Supervision based RE, which require negligible human supervision. Recently, there has been a continuously increasing trend in RE research towards distant supervision based techniques.

Though the state-of-the-art for RE has improved a lot in the last decade, there are still many promising future research directions in RE. We list some of these potential directions below:\\
%\begin{enumerate}
%\item 
1. There have been several techniques for joint modelling of entity and relation extraction. However, the best reported F-measure on ACE 2004 dataset when gold-standard entities are not given, is still very low at around 48\%. This almost 30\% lower than the F-measure achieved when gold-standard entity information is assumed. Hence, there is still some scope of improvement here with more sophisticated models.\\
%\item 
2. There has been little work for extracting n-ary relations, i.e. relations involving more than two entity mentions. There is a scope for more useful and principled approaches for this.\\
%\item 
3. Most of the RE research has been carried out for English, followed by Chinese and Arabic, as ACE program released the datasets for these 3 languages. It would be interesting to analyse how effective and language independent are the existing RE techniques. More systematic study is required for languages with poor resources (lack of good NLP pre-processing tools like POS taggers, parsers) and free word order, e.g. Indian languages.\\
%\item 
4. Depth of the NLP processing used in most of the RE techniques, is mainly limited to lexical and syntax (constituency and dependency parsing) and few techniques use light semantic processing. It would be quite fruitful to analyse whether deeper NLP processing such as semantics and discourse level can help in improving RE performance.
%\end{enumerate}

For new entrants in the field, this survey would be quite useful to get introduced to the RE task. They would also get to know about various types of RE techniques and evaluation methods. This survey is also useful for the practitioners as they can get a quick overview of the RE techniques and decide which technique best suits their specific problem. For researchers in the field, this survey would be useful to get an overview about most of the RE techniques proposed in last decade or so. They can learn about how the techniques evolved over time, what are the pros and cons of each technique and what is the relative performance of these techniques. We have also pointed out some of the recent trends in RE techniques and which would be useful for the researchers. We have also listed some open problems in RE which can lead to new research in future.

%\blindtext
% or


%\appendices
%Some text for the appendix.

\section*{Acknowledgement}
The authors would like to thank Swapnil Hingmire for his efforts of reviewing the draft and providing several useful suggestions for improvement.

\bibliographystyle{plain}
\bibliography{final}

%\begin{biography}[{\includegraphics[width=1in,height=1.25in,clip,keepaspectratio]{mshell}}]{Michael Shell}

%\begin{IEEEbiography}[{\includegraphics[width=1in,height=1.25in,clip,keepaspectratio]{picture}}]{John Doe}
%\blindtext
%\end{IEEEbiography}


%\vfill


\end{document}